\documentclass[10pt,twocolumn,letterpaper]{article}

\usepackage{color}
\definecolor{green}{rgb}{0, 0.5, 0}
\definecolor{orange}{rgb}{0.8, 0.6, 0.2}
\definecolor{red}{rgb}{1.0, 0.0, 0.0}
\definecolor{teal}{rgb}{0.0, 0.4, 0.4}
\definecolor{purple}{rgb}{0.65,0,0.65}
\definecolor{saffron}{rgb}{0.95,0.75,0.2}
\definecolor{turquoise}{rgb}{0.0,0.5,0.5}
\definecolor {mygray}{gray}{.9}

\newcommand{\etal}{et al.}

\usepackage{booktabs}
\usepackage{colortbl}
\usepackage{makecell}
\usepackage{multirow}
\usepackage{multicol}
\usepackage{array}
\usepackage{gensymb}
\newcolumntype{L}[1]{>{\raggedright\arraybackslash}p{#1}}
\newcolumntype{C}[1]{>{\centering\arraybackslash}p{#1}}
\newcolumntype{R}[1]{>{\raggedleft\arraybackslash}p{#1}}
\usepackage{amsmath}

\usepackage{tabu}
\usepackage{graphicx}

\usepackage{iccv}
\usepackage{times}
\usepackage{epsfig}
\usepackage{graphicx}
\usepackage{amsmath}
\usepackage{amssymb}
\usepackage{makecell}
\usepackage{bbding}
\usepackage{pifont}
\usepackage[ruled,linesnumbered]{algorithm2e}

\usepackage[pagebackref=true,breaklinks=true,letterpaper=true,colorlinks,bookmarks=false]{hyperref}

\iccvfinalcopy % *** Uncomment this line for the final submission

\def\iccvPaperID{2618} % *** Enter the ICCV Paper ID here

\ificcvfinal\fi

\begin{document}

%%%%%%%%% TITLE
%\title{EgoVM: Precise Ego-Localization on Vectorized Maps for Autonomous Driving}
%\title{EgoVM: Precise Ego-Localization with Vectorized Maps in Autonomous Driving}
%\title{EgoVM: Leveraging Lightweight Vectorized Maps for Precise Ego-Localization}
%\title{EgoVM: Leveraging Vectorized Maps for Precise and Robust Ego-Localization}
%\title{EgoVM: A Precise and Robust Ego-Localization Network using Vectorized Maps}
\title{EgoVM: Achieving Precise Ego-Localization using Lightweight Vectorized Maps}

\author{Yuzhe He \hspace{0.5cm} Shuang Liang \hspace{0.5cm} Xiaofei Rui \hspace{0.5cm} Chengying Cai \hspace{0.5cm} Guowei Wan\thanks{Author to whom correspondence should be addressed}\\
	Baidu Autonomous Driving Technology Department (ADT)\\
	%Baidu Intelligent Driving Group (IDG)\\
	{\tt\small \{heyuzhe, liangshuang18, ruixiaofei, caichengying, wanguowei\}@baidu.com}
}

\maketitle
\ificcvfinal\thispagestyle{empty}\fi

\begin{abstract}
    %Accurate and reliable ego-localization is critical for autonomous driving. 
    %We present EgoVM, an end-to-end localization network based on lightweight vectorized maps that achieves comparable localization accuracy to prior state-of-the-art methods using heavy point-based maps.
    %To begin with, BEV features are extracted from online multi-view images and LiDAR point cloud.
    %Then, a set of learnable semantic embeddings are employed to encode the semantic types of map elements, supervised by semantic segmentation, to construct a unified feature representation of map elements in consistency with BEV features.
    %After that, map queries, composed of learnable semantic embeddings and geometric coordinates of map elements, are fed into a transformer decoder to perform cross-modality matching with BEV features.
    %Finally, a robust histogram-based pose solver is adopted to estimate the optimal pose by searching exhaustively over candidate poses.
    %We comprehensively validate the effectiveness of our method using both nuScenes dataset and a newly collected dataset.
    %The experimental results show that our method achieves centimeter-level localization accuracy, and is far superior to existing methods using vectorized maps.
    %Furthermore, our model has been extensively tested in a large fleet of autonomous vehicles under various challenging urban scenes.

    Accurate and reliable ego-localization is critical for autonomous driving. 
    In this paper, we present EgoVM, an end-to-end localization network that achieves comparable localization accuracy to prior state-of-the-art methods, but uses lightweight vectorized maps instead of heavy point-based maps. 
    To begin with, we extract BEV features from online multi-view images and LiDAR point cloud. 
    %Then, we employ a set of learnable semantic embeddings to encode the semantic types of map elements, supervised by semantic segmentation, to construct a unified feature representation of map elements that is consistent with BEV features. 
    Then, we employ a set of learnable semantic embeddings to encode the semantic types of map elements and supervise them with semantic segmentation, to make their feature representation consistent with BEV features.
    After that, we feed map queries, composed of learnable semantic embeddings and coordinates of map elements, into a transformer decoder to perform cross-modality matching with BEV features. 
    Finally, we adopt a robust histogram-based pose solver to estimate the optimal pose by searching exhaustively over candidate poses. 
    We comprehensively validate the effectiveness of our method using both the nuScenes dataset and a newly collected dataset. 
    The experimental results show that our method achieves centimeter-level localization accuracy, and outperforms existing methods using vectorized maps by a large margin. 
    Furthermore, our model has been extensively tested in a large fleet of autonomous vehicles under various challenging urban scenes.
\end{abstract}
\section{Introduction}
\label{section:intro}

% my version
%Several studies have examined online high-definition (HD) maps derived from bird's eye view (BEV) models \cite{li2021hdmapnet, liu2022vectormapnet, liao2022MapTR}, but they may have stability and robustness issues due to obstacle occlusions, road abrasion, and inadequate model capabilities, \etc, which cannot meet the high quality standards of fully autonomous driving.
%Consequently, HD maps remain an indispensable component of fully autonomous vehicles since they can provide comprehensive expression and detailed information about road infrastructure.
%To utilize HD maps as priors, centimeter-level ego-localization becomes necessary \cite{Badue2021survey}.
% ChatGPT version
Online high-definition (HD) maps derived from bird’s eye view (BEV) models have been examined by several studies \cite{li2021hdmapnet, liu2022vectormapnet, liao2022MapTR}. 
However, these maps may suffer from stability and robustness issues due to obstacle occlusions, road abrasion, and inadequate model capabilities, \etc, which cannot meet the high quality standards of fully autonomous driving. 
Therefore, HD maps remain an indispensable component of fully autonomous vehicles as they can provide comprehensive and detailed information about road infrastructure. 
To utilize HD maps as priors, centimeter-level ego-localization is necessary \cite{Badue2021survey}.

% 1. LiDAR-based methods, the drawback: map size.
% my version
%Several methods with the aid of 3D light detection and ranging (LiDAR) scanners have achieved this goal \cite{Kummerle2009parking, levinson2007map, levinson2010robust, wolcott2015fast, wolcott2017robust, Kim2017Vertical, Wan2018MSF, liu2019precise}.
%In these methods, localization maps are typically represented as points, voxels, or Gaussian distributions on 2D grids, which require enormous storage on the vehicle.
%It poses significant challenges to deploy the map covering vast areas onto the vehicle.
%Several works propose to shrank the map size by using pole-like objects \cite{2019arXiv191010550S, Barros2020pole}, vertical corners \cite{Im2016vertical}, footprints and surfaces \cite{Javanmardi2019abstract} of buildings.
% ChatGPT version
Several methods have achieved this goal with the aid of 3D light detection and ranging (LiDAR) scanners \cite{Kummerle2009parking, levinson2007map, levinson2010robust, wolcott2015fast, wolcott2017robust, Kim2017Vertical, Wan2018MSF, liu2019precise}. 
In these methods, localization maps are typically represented as points, voxels, or Gaussian distributions on 2D grids, which require enormous storage on the vehicle. 
This poses significant challenges to deploy the map covering vast areas onto the vehicle. 
To reduce the map size, several works propose to use pole-like objects \cite{2019arXiv191010550S, Barros2020pole}, vertical corners \cite{Im2016vertical}, or footprints and surfaces \cite{Javanmardi2019abstract} of buildings as map features.
% 2. camera-based methods, hd maps
% my version
%The majority of camera-based methods \cite{wu2013local, Qu2015VehicleLU, Vivacqua2018local, Xiao2020local, Pauls2020local, Jeong2020local, Guo2021local, Wang2021icralocal, Qin2021icralocal, Cheng2021rallocal, Petek2022local, Wang2022icralocal, Liang2022icralocal} rely on HD maps that include lane lines, road markings, poles, and traffic signs, \etc. 
%Even though these methods have made great strides over the years, their capabilities still lag behind LiDAR-based techniques, so they are rarely used for fully autonomous driving independently.
% ChatGPT version
Most of the camera-based methods \cite{wu2013local, Qu2015VehicleLU, Vivacqua2018local, Xiao2020local, Pauls2020local, Jeong2020local, Guo2021local, Wang2021icralocal, Qin2021icralocal, Cheng2021rallocal, Petek2022local, Wang2022icralocal, Liang2022icralocal} depend on HD maps that contain lane lines, road markings, poles, traffic signs, \etc. 
Despite the significant progress made by these methods over the years, they still fall behind LiDAR-based techniques in terms of performance.
% 3. end-to-end networks
% my version
%In recent years, end-to-end localization networks \cite{Brsan2018LearningTL, Lu2019L3Net, Wei2019compress, Ma2019semantic, zhou2020da4ad} have shown their potential for improving localization accuracy, scene generalization, and map compression.
%Nonetheless, there are some limitations, such as the large map size \cite{Brsan2018LearningTL, Lu2019L3Net}, the model being associated with the map due to its deep features \cite{Wei2019compress, zhou2020da4ad}, and the only suitability for simple scenes \cite{Ma2019semantic}.
% ChatGPT version
End-to-end localization networks \cite{Brsan2018LearningTL, Lu2019L3Net, Wei2019compress, Ma2019semantic, zhou2020da4ad} have demonstrated their potential for enhancing localization accuracy, scene generalization, and map compression in recent years. 
However, these networks also face some limitations, such as the large map size \cite{Brsan2018LearningTL, Lu2019L3Net}, the model-map dependency \cite{Wei2019compress, zhou2020da4ad}, and the only applicability for simple scenes \cite{Ma2019semantic}.

\begin{figure}[!!t]
	\centering
	\includegraphics[width=1.0\linewidth]{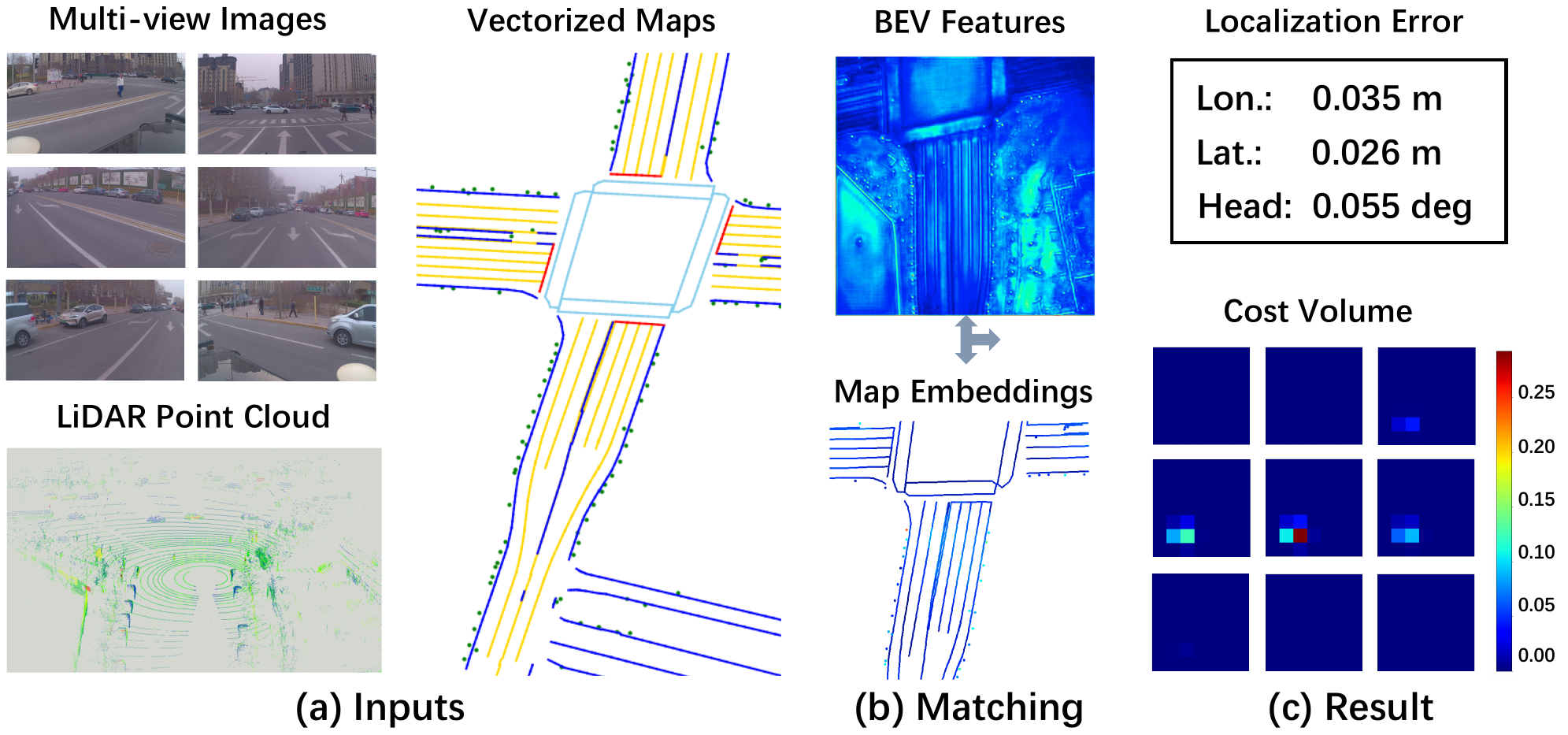}
	\caption{
		% my version
		%The illustration of our method. 
		%(a) The inputs are multi-view images, LiDAR point cloud and vectorized maps. 
		%(b) BEV features are extracted by the BEV feature extraction module, and interacted with vectorized maps to generate map embeddings by the cross-modality matching module.
		%(c) Centimeter-level localization is accomplished by the pose solver module, which takes BEV features and map embeddings as inputs.
		% ChatGPT version
		The illustration of our method. 
		(a) The inputs consist of multi-view images, LiDAR point cloud and vectorized maps. 
		(b) The BEV feature extraction module extracts BEV features, which are then interacted with vectorized maps to generate map embeddings by the cross-modality matching module. 
		(c) The pose solver module takes BEV features and map embeddings as inputs and accomplishes centimeter-level localization.
	}
	\label{fig:teaser}
	\vspace{-0.4cm}
\end{figure}

% a practical localization system:
% 1. high accuracy
% 2. high reliability
% 3. good scalability
% my version
%A practical localization system for fully autonomous driving should have the following characteristics.
%As a first step, \textbf{high accuracy}, which means the horizontal localization error should be within a few centimeters and the heading error should be within a few tenths of a degree.
%The second is \textbf{high reliability}, which is to maintain centimeter-level accuracy in various complex urban scenes.
%The final point is \textbf{good scalability}, which means that the localization map can be mass produced and updated easily.
%Furthermore, the localization map should require less storage to facilitate onboard deployment, and the localization model should be upgradeable without updating the map.
% ChatGPT version
A practical localization system for fully autonomous driving should meet the following requirements.
First, it should achieve \textbf{high accuracy}, with the horizontal localization error and the heading error being within a few centimeters and a few tenths of a degree, respectively.
Second, it should ensure \textbf{high reliability}, with the ability to maintain centimeter-level accuracy in various complex urban scenes.
Third, it should have \textbf{good scalability}, with the capability to generate and update the localization map efficiently and massively.
Moreover, the localization map should have low storage demand to enable onboard deployment, and the model should be upgradeable without updating the map.

To achieve the above goals, we propose a novel ego-localization network, dubbed \textbf{EgoVM} (\textbf{Ego}-localization using \textbf{V}ectorized \textbf{M}aps). 
EgoVM takes multi-view images from surrounding cameras, 3D points from LiDAR sensor, and offline vectorized maps as inputs, and estimates the pose offset relative to the initial pose in an end-to-end manner. 
Figure~\ref{fig:teaser} shows the illustration of our method. 
In our method, we only estimate 2D position and heading offsets of a 6-Dof pose as in \cite{levinson2010robust, wolcott2015fast, Wan2018MSF}. 
Generating BEV features from multi-view images, 3D LiDAR points or both has been applied in 3D object detection \cite{li2022bevformer, Liu2022BEVFusionMM} and semantic segmentation \cite{Peng2022BEVSegFormerBE, zhou2022cross}. 
Therefore, we attempt to perform matching between vectorized maps and BEV features to estimate 3-DoF pose (2D position and heading) offset in BEV space. 
However, they are still different modalities and difficult to compare, even if they are transformed into a canonical view. 
To address this issue, we employ a set of learnable embeddings to describe the \textit{semantic types} of vectorized map elements, which include \textit{lane line}, \textit{pedestrian crossing}, \textit{road marking}, \textit{pole} and \textit{traffic sign}, \etc. 
%The learnable embeddings are supervised by semantic segmentation to construct unified features of map elements that are consistent with BEV features. 
The learnable embeddings are supervised by semantic segmentation to make their features consistent with BEV features. 
Then, vectorized map elements, represented by learnable semantic embeddings and geometric coordinates, interact with BEV features through a transformer decoder for cross-modality matching. 
Finally, we apply a robust histogram-based pose solver \cite{Brsan2018LearningTL, Lu2019L3Net} to determine the optimal pose offset by searching exhaustively over 3-DoF candidate poses.

In summary, our main contributions are:
\vspace{-0.2cm}
\begin{itemize}
	\item An end-to-end localization network that uses light vectorized maps, which achieves centimeter-level localization accuracy comparable to those methods that use heavy point-based maps and is far superior to existing methods that use vectorized maps.
	\vspace{-0.2cm}
	\item A novel design for cross-modality matching, which adopts a set of learnable semantic embeddings and a transformer decoder to bridge the representation gap between vectorized maps and BEV features. 
	\vspace{-0.2cm}
	\item Comprehensive tests and detailed ablation analysis on real-world datasets to verify the effectiveness of the proposed method. 
	\vspace{-0.2cm}
	\item Integration with other sensors (GNSS, IMU) for a multi-sensor fusion localization system that has been extensively tested in various challenging urban scenes.
\end{itemize}

\section{Related Work}
\label{section:related}

%The survey work from A. Chalvatzaras \etal \cite{Chalvatzaras2022survey} provides a detailed overview of map-based localization techniques. A comprehensive discussion of these methods is beyond the scope of this work.

% my version
%\noindent
%\textbf{Localization Using Point-based Maps.}
%Methods of aligning multiple passes of LiDAR point clouds over the same area to construct accurate maps and matching the online sensory input with the map have been widely used in fully autonomous vehicles.
%The pioneering works \cite{levinson2007map, levinson2010robust} represent maps using LiDAR intensities, which could provide texture information about the environment.
%Altitude information is combined with intensities in the following works \cite{Kim2017Vertical, Wan2018MSF} to achieve more robust and accurate localization.
%The work of R. Wolcott \etal \cite{wolcott2015fast, wolcott2017robust} employs Gaussian Mixture Model (GMM) to describe intensity and altitude information, which works well in snowy scenes and has the potential to handle overpass situations.
%H. Liu \etal \cite{liu2019precise} proposes to extract low-level semantic segmentation-based features, including ground, road-curb, surface, and edge.
%These methods have led to compelling performance, but their high storage requirements limit their use.

% ChatGPT version
\noindent
\textbf{Localization Using Point-based Maps.} 
Methods that align multiple passes of LiDAR point clouds over the same area to construct accurate maps and match the online sensory input with the map have been widely used in fully autonomous vehicles. 
The pioneering works \cite{levinson2007map, levinson2010robust} represent maps using LiDAR intensities, which could provide texture information about the environment. 
Subsequent works \cite{Kim2017Vertical, Wan2018MSF} combine altitude information with intensities to achieve more robust and accurate localization. 
The work of R. Wolcott \etal \cite{wolcott2015fast, wolcott2017robust} employs Gaussian Mixture Model (GMM) to describe intensity and altitude information, which works well in snowy scenes and has the potential to handle overpass situations. 
H. Liu \etal \cite{liu2019precise} proposes to extract low-level semantic segmentation-based features, including ground, road-curb, surface, and edge. 
These methods have achieved impressive performance, but their high storage requirements limit their applicability.

% my version
%\vspace{0.1cm}
%\noindent
%\textbf{Localization Using Vectorized Maps.}
%A compact representation is to leverage lightweight vectorized maps, which contain geometric and semantic information of the scene, such as lane line \cite{MarkusSchreiber2013LaneLocLM, DixiaoCui2014RealtimeGL, Vivacqua2018local}, road marking \cite{AnanthRanganathan2013LightweightLF, wu2013local, KichunJo2015PreciseLO, JaeKyuSuhr2017SensorFL, YanLu2017MonocularLI, Qin2021icralocal}, pole \cite{RobertSpangenberg2016PolebasedLF}, traffic sign \cite{Qu2015VehicleLU}, and their combination \cite{YufengYu2014MonocularVL, Pauls2020local, Cheng2021rallocal, Xiao2020local, Guo2021local, Wang2021icralocal}.
%In the case of lane lines and road markings, the matching process is performed either from perspective or from bird's eye view.
%The majority of methods perform matching under perspective view when poles or traffic signs are involved.
%Several recent methods propose using coarse-to-fine strategy \cite{Guo2021local}, distance transform \cite{Pauls2020local}, or robust data association \cite{Wang2021icralocal, Cheng2021rallocal} in order to improve the robustness and availability of localization systems.
%Our method seeks to incorporate vectorized maps to achieve accurate and reliable centimeter-level localization for fully autonomous driving due to their compactness.

% ChatGPT version & little modification
\vspace{0.1cm}
\noindent
\textbf{Localization Using Vectorized Maps.} 
%A compact representation is to leverage lightweight vectorized maps, which contain geometric and semantic information of the scene, such as lane line \cite{MarkusSchreiber2013LaneLocLM, DixiaoCui2014RealtimeGL, Vivacqua2018local}, road marking \cite{AnanthRanganathan2013LightweightLF, wu2013local, KichunJo2015PreciseLO, JaeKyuSuhr2017SensorFL, YanLu2017MonocularLI, Qin2021icralocal}, pole \cite{RobertSpangenberg2016PolebasedLF}, traffic sign \cite{Qu2015VehicleLU}, and their combination \cite{YufengYu2014MonocularVL, Pauls2020local, Cheng2021rallocal, Xiao2020local, Guo2021local, Wang2021icralocal}. 
A compact representation can be achieved by utilizing lightweight vectorized maps, which contain geometric and semantic information of the scene, such as lane lines \cite{MarkusSchreiber2013LaneLocLM, DixiaoCui2014RealtimeGL, Vivacqua2018local}, road markings \cite{AnanthRanganathan2013LightweightLF, wu2013local, KichunJo2015PreciseLO, JaeKyuSuhr2017SensorFL, YanLu2017MonocularLI, Qin2021icralocal}, poles \cite{RobertSpangenberg2016PolebasedLF}, traffic signs \cite{Qu2015VehicleLU}, and their combinations \cite{YufengYu2014MonocularVL, Pauls2020local, Cheng2021rallocal, Xiao2020local, Guo2021local, Wang2021icralocal}.
In the case of lane lines and road markings, the matching process can be performed either from perspective or from bird’s eye view. 
The majority of methods perform matching from perspective view when poles or traffic signs are involved. 
Several recent methods propose using coarse-to-fine strategy \cite{Guo2021local}, distance transform \cite{Pauls2020local}, or robust data association \cite{Wang2021icralocal, Cheng2021rallocal} to improve the robustness and availability of localization systems. 
Our method aims to incorporate vectorized maps to achieve accurate and reliable centimeter-level localization for fully autonomous driving due to their compactness.

% my version
%\vspace{0.1cm}
%\noindent
%\textbf{End-to-end Localization Networks.}
%End-to-end localization networks that compute similarity between online LiDAR sweeps and intensity map \cite{Brsan2018LearningTL} or keypoint map \cite{Lu2019L3Net} have been exploited.
%They coincidentally apply a 3D cost volume to estimate horizontal and heading offsets in an exhaustive searching way.
%A similar solution \cite{zhou2020da4ad}  exploits to search for long-term salient, distinctive, and stable features to achieve centimeter-level visual localization.
%X. Wei \etal \cite{Wei2019compress} proposes to learn to compress the map without loss of localization accuracy.
%Due to deep features residing in the map,  the map must be updated for model upgrades, which does not facilitate map deployment and model iteration.
%W. Ma \etal \cite{Ma2019semantic} exploits lanes and traffic signs to localize against a sparse semantic map that requires orders of magnitude less storage than previous approaches.
%Zhang \etal \cite{zhang2022bev} develops an end-to-end visual localization method based on HDMap and BEV representation.

% ChatGPT version & little modification
\vspace{0.1cm}
\noindent
\textbf{End-to-end Localization Networks.} 
End-to-end localization networks that compute similarity between online LiDAR sweeps and intensity map \cite{Brsan2018LearningTL} or keypoint map \cite{Lu2019L3Net} have been explored. 
They both apply a 3D cost volume to estimate horizontal and heading offsets in an exhaustive searching way. 
A similar solution \cite{zhou2020da4ad} exploits long-term salient, distinctive, and stable features to achieve centimeter-level visual localization. 
X. Wei \etal \cite{Wei2019compress} proposes to learn to compress the map without loss of localization accuracy. 
However, due to deep features residing in the map, the map must be updated for model upgrades, which does not facilitate map deployment and model iteration. 
W. Ma \etal \cite{Ma2019semantic} exploits lanes and traffic signs to localize against a sparse semantic map that requires orders of magnitude less storage than previous approaches. 
Zhang \etal \cite{zhang2022bev} proposes an end-to-end visual localization method based on HD Map and BEV representation, similar but different from ours.

\begin{figure*}[!htbp]
	\centering
	\includegraphics[width=\linewidth]{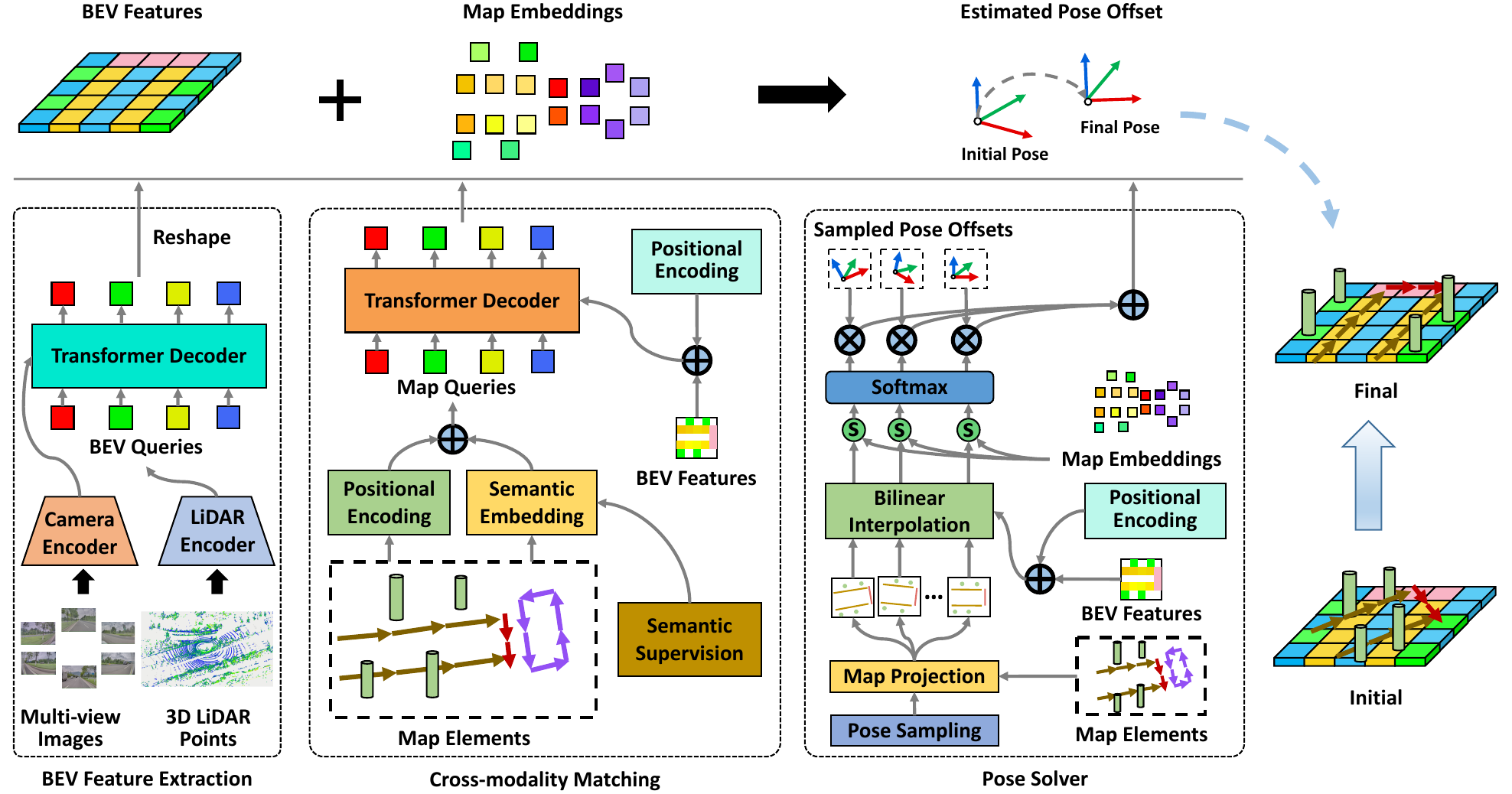}
	\caption{
		% my version
		%The network architecture of EgoVM.
		%Firstly, the multi-view images and 3D LiDAR points are fed into camera encoder and LiDAR encoder respectively, and fused to BEV features.
		%Then, vectorized map elements and BEV features are taken as the query and key / value of the transformer decoder for cross-modality matching, thus obtaining map embeddings.
		%Finally, map embeddings are compared with sampled map features by candidate poses to calculate their similarities, so as to estimate the optimal pose offset.
		% ChatGPT version & little modification
		The network architecture of EgoVM. 
		First, the multi-view images and 3D LiDAR points are fed into camera encoder and LiDAR encoder respectively, and fused to obtain BEV features. 
		Second, vectorized map elements and BEV features are used as the queries and keys / values of the transformer decoder for cross-modality matching, thus generating map embeddings. 
		Third, map embeddings are compared with interpolated map features by candidate poses to calculate their similarities, so as to estimate the optimal pose offset.
	}
	\label{fig:network}
	\vspace{-0.3cm}
\end{figure*}
 
% my version
%\vspace{0.1cm}
%\noindent
%\textbf{BEV Feature Extraction.}
%On account of prediction and planning tasks operating under bird's eye view, lots of methods have tried to generate BEV features from multi-view images and perform perception task.
%One series of methods first performs monocular depth estimation, and then lifts 2D image features to 3D space and splat to BEV \cite{JonahPhilion2022LiftSS, CodyReading2021CategoricalDD, AnthonyHu2021FIERYFI, JunjieHuang2022BEVDetHM, YinhaoLi2022BEVDepthAO}.
%Another series of methods employs transformer \cite{AshishVaswani2017AttentionIA} to perform view transformation \cite{li2022bevformer, WeixiangYang2021ProjectingYV, YingfeiLiu2022PETRv2AU,Peng2022BEVSegFormerBE, zhou2022cross}.
%There are several LiDAR-camera fusion strategies. 
%One is point decoration fusion, which acquires image features or semantic scores for LiDAR points and then generate BEV features, \eg, PointPainting \cite{Vora2020PointPainting}, PointAugmenting \cite{Wang2021PointAugmenting}. 
%A second approach is to fuse the LiDAR and image BEV features directly under the BEV representation, \eg, BEVFusion \cite{Liu2022BEVFusionMM}.
%The last kind adopts transformer to perform multi-modality fusion, \eg, DeepFusion \cite{Li2022DeepFusion}, TransFusion \cite{Bai2022Transfusion}.

% ChatGPT version
\vspace{0.1cm}
\noindent
\textbf{BEV Feature Extraction.} 
Since prediction and planning tasks operate under bird’s eye view, many methods have attempted to generate BEV features from multi-view images and perform perception task. 
One series of methods first performs monocular depth estimation, and then lifts 2D image features to 3D space and splats to BEV \cite{JonahPhilion2022LiftSS, CodyReading2021CategoricalDD, AnthonyHu2021FIERYFI, JunjieHuang2022BEVDetHM, YinhaoLi2022BEVDepthAO}. 
Another series of methods employs transformer \cite{AshishVaswani2017AttentionIA} to perform view transformation \cite{li2022bevformer, WeixiangYang2021ProjectingYV, YingfeiLiu2022PETRv2AU,Peng2022BEVSegFormerBE, zhou2022cross}. 
There are also several LiDAR-camera fusion strategies. 
One is point decoration fusion, which acquires image features or semantic scores for LiDAR points and then generates BEV features, \eg, PointPainting \cite{Vora2020PointPainting}, PointAugmenting \cite{Wang2021PointAugmenting}. 
A second approach is to fuse the LiDAR BEV features and image BEV features directly under the BEV representation, \eg, BEVFusion \cite{Liu2022BEVFusionMM}. 
The last kind adopts transformer to perform multi-modality fusion, \eg, DeepFusion \cite{Li2022DeepFusion}, TransFusion \cite{Bai2022Transfusion}.

\section{Problem Formulation}
\label{sec:problem}

Our goal is to estimate an optimal pose offset given an online point cloud, multi-view images, pre-built vectorized maps, and an initial pose. 
The pre-built maps contain vectorized elements such as lane lines, road boundaries, pedestrian crossings, stop lines, road markings, traffic signs, and poles, denoted as ${ M_i | i=1,2,\cdots, K}$, where $M_i\in \mathbb{R}^{m_i}$ is the $i$-th vectorized map element. 
Specifically, lane line, road boundary and stop line are described as horizontal segment on the BEV plane, expressed as endpoints $(x_s, y_s, x_e, y_e)\in \mathbb{R}^4$. 
The pedestrian crossing is represented by segments of adjacent endpoints of a polygon with $(x_i,y_i, x_{i+1},y_{i+1})\in \mathbb{R}^4$. 
The traffic sign and pole are vectorized as ${(x,y,0,h)}\in \mathbb{R}^4$, where $(x,y)$ and $h$ are the center point and height, respectively. 
Our model also takes an initial pose as an input, which can be provided by a multi-sensor fusion localization system. 
The estimated pose offset only consists of the 2D horizontal and heading offsets represented as $\Delta T = (\Delta x, \Delta y, \Delta \psi)$, following classical LiDAR-based localization methods \cite{levinson2010robust, wolcott2015fast, Wan2018MSF}.
\section{Method}
\label{section:method}

%\input{sources/method/problem}
%\subsection{Overall Architecture}

% my version
%As shown in Figure~\ref{fig:network}, EgoVM can be conceptually divided into three parts.
%The multi-view image features and BEV LiDAR features are first extracted, and then fused into unified BEV features (Section~\ref{subsec:bev}).
%Then, a set of learnable embeddings encoding map element types, supervised by semantic segmentation, interact with BEV features by a transformer decoder to perform cross-modality matching, thus obtaining map embeddings (Section~\ref{subsec:cross}).
%Finally, the pose solver samples several candidate poses to project map elements to BEV plane, obtaining corresponding features by bilinear interpolation, then compares them with map embeddings to estimate the optimal pose offset (Section~\ref{subsec:pose}).

% ChatGPT version 
Figure~\ref{fig:network} shows the three parts of EgoVM. 
First, the multi-view image features and LiDAR BEV features are extracted and fused into unified BEV features (Section~\ref{subsec:bev}). 
Second, a set of learnable embeddings encoding map element types, supervised by semantic segmentation, interact with BEV features through a transformer decoder to perform cross-modality matching, thus obtaining map embeddings (Section~\ref{subsec:cross}). Third, the pose solver samples several candidate poses to project map elements to BEV plane, obtaining corresponding features by bilinear interpolation, and then compares them with map embeddings to estimate the optimal pose offset (Section~\ref{subsec:pose}).
\subsection{BEV Feature Extraction}
\label{subsec:bev}
We adopt a transformer decoder for fusing image features and LiDAR BEV features. 
First, multi-view images and LiDAR points are fed into a camera encoder and a LiDAR encoder to extract image features and LiDAR BEV features respectively. 
Then, the transformer decoder takes LiDAR BEV features to initialize BEV queries and interact with image features, thus obtaining fused BEV features.

\vspace{0.1cm}
\noindent
\textbf{Camera Encoder.}
The multi-view images $\{I_i|i=1,2,\cdots,N\}$ are fed into a shared backbone network (\eg, ResNet\cite{He2016DeepRL}, VoVNet\cite{Lee2020CenterMaskRA}), followed by FPN \cite{lin2017fpn} to extract multi-scale features, denoted as $F^I=\{F^I_{ij}|i=1,2,\cdots,N,j=1,2,\cdots,L\}$, where $F^I_{ij}\in \mathbb{R}^{H_j\times W_j \times C}$ is the $j$-th level feature of $i$-th camera, and $H_j,W_j$ are the height and width of $j$-th level feature respectively.

\vspace{0.1cm}
\noindent
\textbf{LiDAR Encoder.}
The 3D LiDAR points $\{P_i|i=1,2,\cdots,n\}$ are first fed into a pillar-based feature extractor (\eg, Pillar Feature Net of PointPillars \cite{Lang2019PointPillarsFE}) to extract pseudo image features. Then, a group of 2D convolutional layers is applied to obtain LiDAR BEV features $F^L\in \mathbb{R}^{H\times W \times C}$ from the pseudo image features, where $H$ and $W$ are the height and width of BEV space.

\vspace{0.1cm}
\noindent
\textbf{BEV Fusion.}
%To generate unified LiDAR-camera BEV features, we adopt a transformer decoder to fuse LiDAR BEV features and multi-view image features based on BEVFormer \cite{li2022bevformer}.
%Specifically, the LiDAR BEV features is used to initialize BEV queries, self-attention is performed on BEV queries, and then cross-attention is applied to aggregate multi-view image features.
%The self-attention and cross-attention layers are implemented based on deformable attention \cite{Zhu2021DeformableDD} for efficiency.
%The cross-attention (CA) layer can be formulated as:
%\begin{equation}
%    \small
%    \mathrm{CA}(F^L_{x,y},F^I)=\frac{1}{N_\mathrm{hit}}\sum_{i\in \mathcal{H}}\sum_{j=1}^{N_\mathrm{ref}} \mathrm{DA}(g(F^L_{x,y}),\mathcal P(p_j,i),F^I_{i}),
%\end{equation}
%where$F^L_{x,y}$ is the BEV query of pillar $(x,y)$, $F^I$ represents the multi-view features, $N_\mathrm{ref}$ is the total reference points for each pillar, $\mathcal{H}$ is the camera indices which are hit by at least one of the reference points, $N_\mathrm{hit}$ is the total hit number, $\mathrm{DA}$ refers to multi-scale deformable attention, $g$ implies the transformation module, and $\mathcal P(p_j,i)$ projects the reference point $p_j$ to the $i$-th camera.
%The fused BEV features are denoted as $F^B \in \mathbb{R}^{H \times W \times C}$.
To generate unified LiDAR-camera BEV features, we adopt a transformer decoder to fuse LiDAR BEV features and multi-view image features based on BEVFormer \cite{li2022bevformer}. 
Specifically, we use LiDAR BEV features to initialize BEV queries, perform self-attention on them, and then apply cross-attention to aggregate multi-view image features. 
The self-attention and cross-attention layers are implemented based on deformable attention \cite{Zhu2021DeformableDD} for efficiency. 
The fused BEV features are denoted as $F^B \in \mathbb{R}^{H \times W \times C}$.

\subsection{Cross-modality Matching}
\label{subsec:cross}
Vectorized map elements differ significantly from BEV features in terms of representation.
To match them, we employ a set of learnable embeddings and a transformer decoder to bridge the representation gap.
The learnable embeddings encode the semantic types of the map elements and then act as queries for the transformer decoder to interact with BEV features and generate unified features for map elements, denoted as \textit{map embeddings}.

\vspace{0.1cm}
\noindent
\textbf{Semantic Embedding.}
Vectorized map elements have different semantic types, such as \textit{lane line}, \textit{road boundary}, \textit{stop line}, \textit{pedestrian crossing}, \textit{road marking}, \textit{pole} and \textit{traffic sign}.
We use a set of learnable embeddings $E^{sem}=\{E^{sem}_j  \in \mathbb R^C | j=1, 2, \cdots, N_e\}$, each of which is expected to learn a specific representation of the corresponding semantic type.
Each map element $M_i$ has a semantic type of $s_i \in \{1, 2, \cdots, N_e\}$ and a correspondence semantic embedding is $E_{s_i}^{sem} \in E^{sem}$.

\vspace{0.1cm}
\noindent
\textbf{Positional Encoding.}
Vectorized map elements $\{ M_i | i=1,2,\cdots, K\}$ are commonly represented in a global coordinate system, such as Universal Transverse Mercator (UTM) coordinate. 
First, we normalize them by: 
\begin{equation} 
    \widehat{M_i} = (M_i - O_{xy}) / R_{xy}, 
\end{equation} 
where $O_{xy}$ is the initial pose coordinates, and $R_{xy}$ is the height and width ranges of the BEV space.

Then, we feed the normalized map elements $\{\widehat M_i \}$ into a shared Multi-Layer Perceptron (MLP) layer to obtain positional encodings $E^{pos} = \{E^{pos}_i \in \mathbb R^C | i=1,2,\cdots, K\ \}$. 

\vspace{0.1cm}
\noindent
\textbf{Matching.}
We apply a transformer decoder to match map elements to BEV features, resulting in \textit{map embeddings} $M^{emb}=\{M_i^{emb} | i = 1, 2, \cdots, K\}$. We initialize the map query $Q_i \in \mathbb R^C(1\leq i\leq K)$ of the map element $M_i$ by adding its semantic embedding and positional encoding:
\begin{equation}
    Q_i = E^{pos}_i + E^{sem}_{s_i}.
\end{equation}

The self-attention module of the transformer decoder is formulated as:
\begin{equation}
    \mathrm{SA}(Q_i) = \sum_{m=1}^M W_m \sum_{i^{\prime}=1}^K A_{m}(Q_i,Q_{i^{\prime}}) W_m^{\prime} Q_{i^{\prime}},
\end{equation}
where $M$ is the number of heads, $W_m$ and $W_m^{\prime}$ are learnable projection matrices, $A_{m}(Q_i,Q_{i^{\prime}})$ is attention weight between map queries $Q_i$ and $Q_{i^{\prime}}$. The cross-attention module is defined as:
\begin{equation}
    \mathrm{CA}(Q_i, F^B) = \mathrm{DA}(Q_i, {r^B_i}, F^B+B^{pos}),
\end{equation}
where DA represents deformable attention, $r^{B}_i\in \mathbb R^2$ is a reference point that is acquired by projecting the endpoint of map element $M_i$ to BEV space via the inital pose, and $B^{pos}$ is the 2D positional encodings of BEV space.

\vspace{0.1cm}
\noindent
\textbf{Semantic Supervision.}
To better learn the semantic embeddings $E^{sem}$, we use an auxiliary network that performs semantic segmentation.
We take the $j$-th semantic type as example.
The semantic probabilities $S_j \in \mathbb{R}^{H \times W}$ of $j$-th semantic type with given BEV features $F^B$ is formulated as:
\begin{equation}
    S_j(h, w) = \mathrm{sigmoid}\left(F^B(h,w) \odot E^{sem}_j\right),
    \label{semsup}
\end{equation}
where $h$ and $w$ are the indices of BEV grid, and $\odot$ represents dot product.

%The ground truth semantic probabilities for $j$-th semantic type can be generated as follows.
%The map elements $\{M_i | s_i = j\}$ are first projected to BEV plane from the global coordinate using the ground truth pose.
%Then, the projected BEV plane is discretized into a $H \times W$ grid with 0 or 1 according to occupancy, which represents the ground truth semantic probabilities of $j$-th semantic type and denoted as $S^{GT}_j \in \mathbb{R}^{H \times W}$.

To generate the ground truth semantic probabilities for each semantic type $j$, we do the following steps. 
First, we project the map elements $\{M_i | s_i = j\}$ to the BEV plane using the ground truth pose. 
Second, we divide the BEV plane into a grid of size $H \times W$ and assign each cell a value of 0 or 1 depending on whether it is occupied by a map element or not. 
This gives us a binary matrix $S^{GT}_j \in \mathbb{R}^{H \times W}$ that represents the ground truth semantic probabilities for semantic type $j$.

\subsection{Pose Solver}
\label{subsec:pose}
Following \cite{Lu2019L3Net}, we utilize a histogram-based pose solver to estimate the optimal pose offset.
%First, several candidate poses are sampled and map elements are projected to BEV plane by sampled poses.
%Then, the probabilities of candidate poses are calculated and the optimal pose is obtained as the expectation of candidate poses.

\vspace{0.1cm}
\noindent
\textbf{Candidate Poses.}
%The candidate pose offsets are sampled along x, y and yaw dimensions by grid searching, denoted as $\{\Delta T_{pqr}=(\Delta x_p, \Delta y_q, \Delta \psi_r)|1\leq p,q,r\leq N_s \}$, and the candidate poses $\{T_{pqr}\}$ are obtained by composing the initial pose and sampled pose offsets.
We sample candidate pose offsets along x, y and yaw dimensions by grid searching, denoted as $\{\Delta T_{pqr}=(\Delta x_p, \Delta y_q, \Delta \psi_r)|1\leq p,q,r\leq N_s \}$, and then generate the candidate poses ${T_{pqr}}$ by composing the initial pose and them.

%\vspace{0.1cm}
%\noindent
%\textbf{Optimal Pose Offset.}
%Each map element $M_i(1\leq i\leq K)$ is projected into BEV plane by these candidate poses $\{T_{pqr}\}$ to obtain BEV features $\{M_i^{bev}(T_{pqr})\}$ by bilinear interpolation.
%The similarity score of a specific candidate pose $T_{pqr}$ is calculated by:
%\begin{equation}
%    S(T_{pqr})=\sum_{i=1}^K h(M_i^{bev}(T_{pqr}) \odot M_i^{emb}),
%    \label{score}
%\end{equation}
%where $M_i^{emb}$ represents the map embedding of $M_i$, and $h$ is a shared MLP.
%The similarity scores of all candidate poses are normalized by \textit{softmax} operation to obtain the probabilities, denoted as $\{P_{pqr}\}$.
%Finally, the estimated pose offset $\Delta T$ and covariance $\Sigma$ are as follows:
%\begin{equation}
%    \Delta T = \sum_{1\leq p,q,r\leq N_s} P_{pqr}  \Delta T_{pqr}.
%\end{equation}
%\begin{equation}
%    \Sigma = \sum_{1\leq p, q, r \leq N_s} P_{pqr}  (\Delta T_{pqr} - \Delta T)(\Delta T_{pqr} - \Delta T)^{T}.
%\end{equation}

\vspace{0.1cm}
\noindent
\textbf{Optimal Pose Offset.}
We project map elements $\{M_i | 1\leq i\leq K\}$ into the BEV plane using a specific candidate pose $T_{pqr}$ to obtain their BEV features $\{M_i^{bev}(T_{pqr}) | 1\leq i\leq K \}$ by bilinear interpolation on fused BEV features $F^B$. 
Then we calculate the similarity score of $F^B$ and map embeddings $M^{emb}$ under candidate pose $T_{pqr}$ by:
\begin{equation}
    S(T_{pqr})=\frac{1}{K}\sum_{i=1}^K h(M_i^{bev}(T_{pqr}) \odot M_i^{emb}),
    \label{score}
\end{equation}
where $M_i^{emb}\in M^{emb}$ is the map embedding of $M_i$, and $h$ is a shared MLP. 
Then, we normalize the similarity scores $\{S(T_{pqr}) | 1\leq p,q,r\leq N_s \}$ of all candidate poses by \textit{softmax} to obtain the posterior probability $p(T_{pqr}|X)$, where $X=\{F^B, M^{emb}\}$.
Finally, we estimate pose offset $\Delta T=E(T|X)$ and covariance $\Sigma=Var(T|X)$ as follows:
\begin{equation} 
    \Delta T = \sum_{1\leq p,q,r\leq N_s} p(T_{pqr}|X) \Delta T_{pqr}. 
\end{equation} 
\begin{equation} 
    \Sigma = \sum_{1\leq p, q, r \leq N_s} p(T_{pqr}|X) (\Delta T_{pqr} - \Delta T)(\Delta T_{pqr} - \Delta T)^{T}. 
\end{equation}

\subsection{Loss Function}
\label{subsec:loss}
%The loss functions consist of three losses for pose offset, and one loss for semantic segmentation.
%We use three losses for pose offset and one loss for semantic segmentation.

\vspace{0.1cm}
\noindent
\textbf{RMSE Loss.}
We define the first loss as the root mean square error (RMSE) between predicted pose offset $\Delta T$ and ground truth offset $\Delta T_{gt}$:
\begin{equation}
    \mathcal L_{rmse}=\|\Lambda^{\frac{1}{2}}U^{T}(\Delta T-\Delta T_{gt})\|_2,
\end{equation}
where $\Sigma=USU^{T}$, and $\Lambda \in \mathbb R^{3\times 3}$ is a diagonal matrix obtained by normalizing the diagonal elements of $S^{-1}$.

%\vspace{0.1cm}
%\noindent
%\textbf{Pose Solver KL loss.}
%The second loss is derived from Kullback-Leibler (KL) divergence $D_{KL}(t(T)||p(T|X))$, where $t(T)$ is the target pose distribution, $X$ denotes the set of BEV features and map embeddings, $p(T|X)$ is the posterior pose distribution.
%\liangshuang{The aim of this loss is to narrow the pose ambiguity of posterior pose distribution to avoid mutiple modes.}
%\liangshuang{After dropping constant terms of KL divergence, the KL loss can be obtained}:
%\begin{equation}
%    \mathcal L_{KL}=-\int t(T)\log p(X|T)dT + \log \int p(X|T)dT,
%    \label{KL}
%\end{equation}
%where $p(X|T)$ is the likelihood function.

\vspace{0.1cm}
\noindent
\textbf{Pose Solver KL loss.}
The second loss is derived from Kullback-Leibler (KL) divergence $D_{KL}(t(T)||p(T|X))$, which aims to regularize the posterior probability distribution. 
After dropping constant terms of KL divergence, the KL loss can be obtained:
\begin{equation}
    \mathcal L_{KL}=-\int t(T)\log p(X|T)dT + \log \int p(X|T)dT,
    \label{KL}
\end{equation}
where $t(T)$ and $p(X|T)$ denote the target probability distribution and likelihood function. Note $p(X|T)\propto p(T|X)$, which is defined in the Section~\ref{subsec:pose}.

Following \cite{Chen2022EProPnPGE}, we define $t(T)=\delta(T-T_{gt})$, and $\delta(\cdot)$ is the Dirac delta function. Equation~\ref{KL} is rewritten as: 
\begin{equation} 
    \mathcal L_{KL}=-\log\frac{\exp(S(T_{gt}))}{\int \exp(S(T))dT}. 
    \label{KL2} 
\end{equation} 

Based on this, the pose solver KL loss is obtained by Monte Carlo integration: 
\begin{equation} 
    \mathcal L^{ps}_{KL}=-\log \frac{\exp(S(T_{gt}))}{\sum_{1\leq p,q,r\leq N_s}\exp(S(T_{pqr}))}. 
\end{equation}

\vspace{0.1cm}
\noindent
\textbf{Random Pose KL Loss.} 
We use a random pose sampling strategy to enhance the supervision further. The sampled poses $T_{j}(1\leq j\leq N_r)$ are drawn from a pose distribution $q(T)$, and the KL loss is calculated as follows: 
\begin{equation} 
    \mathcal L^{rp}_{KL}=-\log \frac{\exp(S(T_{gt}))}{\frac{1}{N_r}\sum_{1\leq j\leq N_r} \frac{1}{q(T_{j})}\exp(S(T_{j}))}, 
\end{equation} 
where $q(T)$ is a combination of a 2-DoF multivariate t-distribution on x and y dimensions, and a mixture of von Mises and uniform distribution on yaw dimension.

\vspace{0.1cm}
\noindent
\textbf{Semantic Segmentation Loss}. 
We use a semantic segmentation loss function that better supervises semantic embeddings and BEV features by summing up the focal losses (FC) of all semantic classes:
\begin{equation}
    \mathcal L_{ss}=\sum_{j=1}^{N_e} \mathrm{FC}(S_j,S_j^{GT}).
\end{equation}

\begin{table*}[h]
    \footnotesize
    \begin{center}
        \scalebox{0.87}{
            \begin{tabular}{c|ccc|ccc|ccc}
                \toprule
                \multirow{2}{*}{Method} &
                \multicolumn{3}{c|}{Longitudinal Error} & \multicolumn{3}{c|}{Lateral Error} & \multicolumn{3}{c}{Yaw Error} \\
                & MAE(m) & RMSE(m) & $<$0.1m/0.2m/0.3m(\%) & MAE(m) & RMSE(m) & $<$0.1m/0.2m/0.3m(\%) & MAE($^{\circ}$) & RMSE($^{\circ}$) & $<$0.1$^{\circ}$/0.3$^{\circ}$/0.6$^{\circ}$(\%) \\

                \midrule
                MSF-LiDAR
                & \underline{0.041} & \textbf{0.052} & \underline{94.31}/\textbf{99.86}/\textbf{99.99}
                & \underline{0.045} & \underline{0.058} & \underline{91.22}/\underline{99.84}/\textbf{99.99}
                & \underline{0.092} & \underline{0.122} & \underline{64.02}/\underline{97.22}/\textbf{99.96} \\
                
                DA4AD
                & 0.089 & 0.200 & 76.20/94.14/96.89 
                & 0.085 & 0.167 & 74.85/93.95/97.06 
                & 0.137 & 0.231 & 53.56/92.76/98.14 \\

                Structure-based
                & 0.318 & 0.372 & 14.66/31.26/49.72
                & 0.143 & 0.191 & 47.85/74.51/87.74
                & 0.206 & 0.267 & 35.33/72.15/97.92 \\

                \midrule
                Ours (Visual)
                & 0.149 & 0.241 & 45.91/77.43/91.30
                & 0.073 & 0.098 & 72.94/95.85/99.52
                & 0.168 & 0.215 & 35.75/86.11/99.02 \\

                % Ours (LiDAR)
                % & 0.037 & 0.063 & 95.72/99.52/99.74
                % & 0.041 & 0.053 & 94.28/98.89/99.99
                % & 0.094 & 0.116 & 58.10/99.24/99.99 \\                

                Ours
                & \textbf{0.035} & \underline{0.087} & \textbf{96.94}/\underline{99.70}/\underline{99.78}
                & \textbf{0.033} & \textbf{0.043} & \textbf{97.44}/\textbf{99.93}/\textbf{99.99}
                & \textbf{0.080} & \textbf{0.106} & \textbf{70.27}/\textbf{98.91}/\underline{99.89} \\

				\bottomrule
            \end{tabular}
        }
    \end{center}
    \vspace{-2mm}
    \caption{
        \small 
        Comparison of localization accuracy with other methods on the self-collected dataset in terms of longitudinal, lateral and heading errors. The best and second best results are highlighted in bold and underline, respectively.
    }
    \label{tab:self_dataset}
	\vspace{-4mm}
\end{table*}
\section{Map Extension}
\label{extension}

%In vectorized maps, lane lines, road markings, stop lines, and pedestrian crossings are appearance features, while poles and traffic signs represent geometric features.
%The localization based solely on appearance features will deteriorate under poor lighting conditions.
%Geometric features, such as poles and traffic signs, contribute to improving the situation.
%However, due to their sparsity in the scene, they can't be applied in all road sections.
%To enhance geometric features in map, we introduce planar features abundant in the scene, called \textit{surfels} \tofix{[]}. 
Vectorized maps consist of appearance features, such as lane lines, road markings, stop lines, and pedestrian crossings, and geometric features, such as poles and traffic signs.
Localization that relies only on appearance features is prone to degradation in low-light conditions.
Geometric features can help to improve the localization performance, but they are sparse and not available in every road section.
Therefore, we propose to use \textit{surfels} \cite{pfister2000surfels}\cite{behley2018efficient}, which are planar features that are rich in the scene, to enhance the geometric features in the map.
%The surfels are represented with $\{P_i=(p,n,r)\in R^7\}$, where $p\in R^2$ is the center point, $n\in R^3$ is the norm vector, $r=(\lambda_1/\lambda_2,\lambda_1/\lambda_3)$, and $\lambda_1<\lambda_2<\lambda_3$ are eigenvalues of covariance matrices of surfels.
%Although surfels are rich in the scene and useful for localization, it is time-consuming to process all available surfels.
%As a result, we filter out some surfels based on eigenvalues and grid sampling, where surfels with $\lambda_1/\lambda_2>0.1$ are first removed, and then only one surfel with minimum $\lambda_1/\lambda_2$ is kept in each grid with resolution of 1m.
%Similar to above vectorized map elements, surfels first interact with BEV features in cross-modality matching module and then are used to predict pose in pose solver module.
We represent surfels as $\{P_i=(p,n,r)\in \mathbb{R}^7\}$, where $p\in \mathbb{R}^2$ is the center point, $n\in \mathbb{R}^3$ is the norm vector, $r=(\lambda_1/\lambda_2,\lambda_1/\lambda_3)$, and $\lambda_1<\lambda_2<\lambda_3$ are the eigenvalues of the covariance matrices of surfels.
Surfels are abundant and beneficial for localization, but processing all of them is inefficient.
Therefore, we apply eigenvalue and grid sampling filters to reduce the number of surfels, by discarding surfels with $\lambda_1/\lambda_2>0.1$ and retaining only one surfel with the smallest $\lambda_1/\lambda_2$ in each 1m grid.
Like the other vectorized map elements, surfels are involved in the cross-modality matching module to interact with BEV features, and in the pose solver module to estimate the pose.

%\liangshuang{
%The surfels are represented with $\{P_i=(p,n,r)\in R^7\}$, whrere $p\in R^2$ is the center point, $n\in R^3$ is the norm vector, $r=(\lambda_1/\lambda_2,\lambda_1/\lambda_3)$, and $\lambda_1<\lambda_2<\lambda_3$ are eigenvalues of covariance matrixes of surfels.
%Similar to above vectorized map elements, surfels first interact with BEV features in cross-modality matching module and then are used to predict pose in pose solver module.
%In particular, we want to reduce the influence of unreliable surfels on localization, such as surfels on dynamic objects.
%To this end, we first build a surfel heatmap from BEV features via a convolutional block, which implies reliability scores of different areas.
%Then we project surfels into the heatmap and calculate the reliability of surfels through bilinear interpolation.
%Finally, the surfels are weighted by reliabilities when predicting the pose offset in the pose solver module.
%In practical, heatmap is used to select reliable surfels offline after training, and it is not used during testing.
%}
\section{Experiments}
\label{section:exp}

\subsection{Datasets}
% my version
%We evaluate the performance of the proposed network on nuScenes and a newly self-collected dataset.
%NuScenes dataset contains more than 28,000 and 6,000 frames in the training and validation datasets, respectively.
%We extract map elements including road boundaries, diverders and pedestrian crossings as in \cite{liu2022vectormapnet}, and use 6 camera images, a LiDAR point cloud, intrinsics and extrinsics, and extracted map elements as inputs of our method.
%We also collected a new dataset that contains 203,645 and 81,877 frames in the training and validation datasets.
%Each frame consists of time-aligned six camera images and one LiDAR point cloud, intrinsics and extrinsics, ground truth pose and surrounding map elements.

% ChatGPT version
We evaluate the performance of the proposed network on two datasets: nuScenes and a newly self-collected dataset. 
The nuScenes dataset contains more than 28,000 frames for training and 6,000 frames for validation. 
%We extract map elements such as road boundaries, dividers and pedestrian crossings following \cite{liu2022vectormapnet}, and use 6 camera images, a LiDAR point cloud, intrinsics and extrinsics, and extracted map elements as inputs of our method. 
The self-collected dataset consists of 203,645 frames for training and 81,877 frames for validation. 
Each frame includes six camera images and one LiDAR point cloud, which are time-aligned, as well as intrinsics and extrinsics, ground truth pose and surrounding map elements.

\subsection{Performance}
% shuang's version
%\noindent
%\textbf{Comparison Methods.}
%Our method is compared with some state-of-the-art localization methods, including MSF-LiDAR \cite{Wan2018MSF}, DA4AD \cite{zhou2020da4ad}, a structure-based method and BEV-Locator \cite{zhang2022bev}.
%MSF-LiDAR is a LiDAR-based localization method with multiple Gaussian distributions.
%DA4AD is an end-to-end visual localization method using a dense 3D feature point map.
%The structure-based method is a visual localization method by matching the 3D landmarks in the HD Map and the perceived 2D landmarks in the online images.  
%BEV-Locator is an end-to-end visual localization method similar to us. However, our method differs from it in many ways such as map query type, semantic supervision, pose solver and so on.
%Our method includes 2 modes: visual and full. where visual mode only uses camera and the full mode uses both camera and LiDAR.

% ChatGPT version & little modification
\noindent
\textbf{Comparison Methods.}
We compare our method with several state-of-the-art methods, namely MSF-LiDAR \cite{Wan2018MSF}, DA4AD \cite{zhou2020da4ad}, a structure-based method and BEV-Locator \cite{zhang2022bev}. 
MSF-LiDAR is a LiDAR-based localization method that models the map as Gaussian distributions on 2D BEV grids. 
DA4AD is an end-to-end visual localization method that leverages a dense 3D feature point map as the prior. 
The structure-based method is a visual localization method that matches the 3D landmarks in the HD Map with the perceived 2D landmarks in the online images.
BEV-Locator is an end-to-end visual localization method that shares some similarities with ours. 
However, our method differs from it in several aspects such as map query type, semantic supervision, pose solver and so on. 
Our method has two modes: visual and full. The visual mode only uses camera images as the input, while the full mode uses both camera images and LiDAR point cloud.

\vspace{0.1cm}
\noindent
\textbf{Self-Collected Dataset.}
We conduct experiments on the self-collected dataset and compare our method with MSF-LiDAR, DA4AD and the structure-based method. 
The comparison results are shown in Table \ref{tab:self_dataset}. 
The localization accuracy is measured by mean absolute error (MAE), root mean square error (RMSE) and the percentage of localization error within a certain threshold (e.g., 0.1m/0.1$^{\circ}$). 
Our method achieves the best MAEs of 0.035m, 0.033m and 0.080$^{\circ}$ in the longitudinal, lateral and yaw dimensions, respectively, which are superior to the other methods, and the RMSEs of our method are also better than the other methods in the lateral and yaw dimensions. 
In addition, the percentages of longitudinal, lateral and yaw errors less than 0.3m/0.3m/0.6$^{\circ}$ are above 99.7\%, which indicates the high stability of our method. 
Compared with DA4AD, the visual mode of our method achieves higher accuracy in the lateral dimension and similar performance in the yaw term. 
Although the structure-based method and our visual mode both take HD Map and images as inputs, the performance of our visual mode surpasses that of the structure-based method in all evaluation metrics. 
However, we also notice that the percentage of longitudinal error less than 0.3m of our method is 99.78\%, which is lower than that of MSF-LiDAR and the visual mode of our method is also inferior to DA4AD in the longitudinal dimension. 
The reason is that in some scenarios, there are few landmarks that provide longitudinal constraints in the map, such as poles and signs, as well as surfels, and then our method degrades in the longitudinal dimension.  
We intend to further explore some new geometric and texture features in our future work.

% shuang's version
%\vspace{0.1cm}
%\noindent
%\textbf{NuScenes Dataset.}
%Table.\ref{tab:nuscenes} presents the comparison results of our visual method, full method and BEV-Locator on the nuScenes dataset.
%The visual mode of our method achieves 0.151m, 0.049m, and 0.094$^{\circ}$ in the longitudinal, lateral and yaw directions, which are less than those of BEV-Locator;
%in addition, our full method achieves higher localization accracy with the help of LiDAR point cloud.

% ChatGPT version & modification
\vspace{0.1cm}
\noindent
\textbf{NuScenes Dataset.}
As shown in Table \ref{tab:nuscenes}, we compare our visual mode, full mode and BEV-Locator on the nuScenes dataset. 
The visual mode of our method achieves lower errors of 0.151m, 0.047m, and 0.092$^{\circ}$ in the longitudinal, lateral and yaw dimensions, respectively, than those of BEV-Locator, which demonstrates the effectiveness of our visual localization method based on HD Map and images. 
Furthermore, our full mode improves the localization accuracy by incorporating LiDAR point cloud.

\begin{table}[h]
    \normalsize
    \begin{center}
        \resizebox{0.48\textwidth}{!}{
            \begin{tabular}{c|cc|cc|cc}
                \toprule
                \multirow{2}{*}{Method} &
                \multicolumn{2}{c|}{Longitudinal Error} & \multicolumn{2}{c|}{Lateral Error} & \multicolumn{2}{c}{Yaw Error} \\
                & MAE(m) & $<$0.3m(\%) & MAE(m) & $<$0.3m(\%) & MAE($^{\circ}$) & $<$$0.6^{\circ}(\%)$ \\
                \midrule
                BEV-Locator
                & 0.178 & -
                & 0.076 & -
                & 0.510 & - \\ 
                % \midrule
                Ours (Visual)
                & 0.151 & 86.96
                & 0.047 & 99.64
                & 0.092 & 99.46 \\
                % \midrule
                Ours 
                & \textbf{0.109} & \textbf{94.55}
                & \textbf{0.034} & \textbf{99.86}
                & \textbf{0.089} & \textbf{99.50} \\
                \bottomrule
            \end{tabular}
        }
    \end{center}
    \vspace{-2mm}
    \caption{\small Accuracy evaluation results on the nuScenes dataset.}
    \label{tab:nuscenes}
    \vspace{-4mm}
\end{table}
\begin{table}[h]
    \footnotesize
    \begin{center}
        \resizebox{0.38\textwidth}{!}{
            \begin{tabular}{cccc}
                \toprule
                Localization Map Size & MSF-LiDAR & DA4AD & Ours \\
                \midrule
                MB/km & 8.36  & 5.92 & \textbf{0.35} \\
                \bottomrule
            \end{tabular}
        }
    \end{center}
    \vspace{-2mm}
    \caption{\small The map sizes of comparison methods.}
    \label{tab:map_size}
    \vspace{-4mm}
\end{table}
\begin{figure*}[!htbp]
	\centering
	\includegraphics[width=\linewidth]{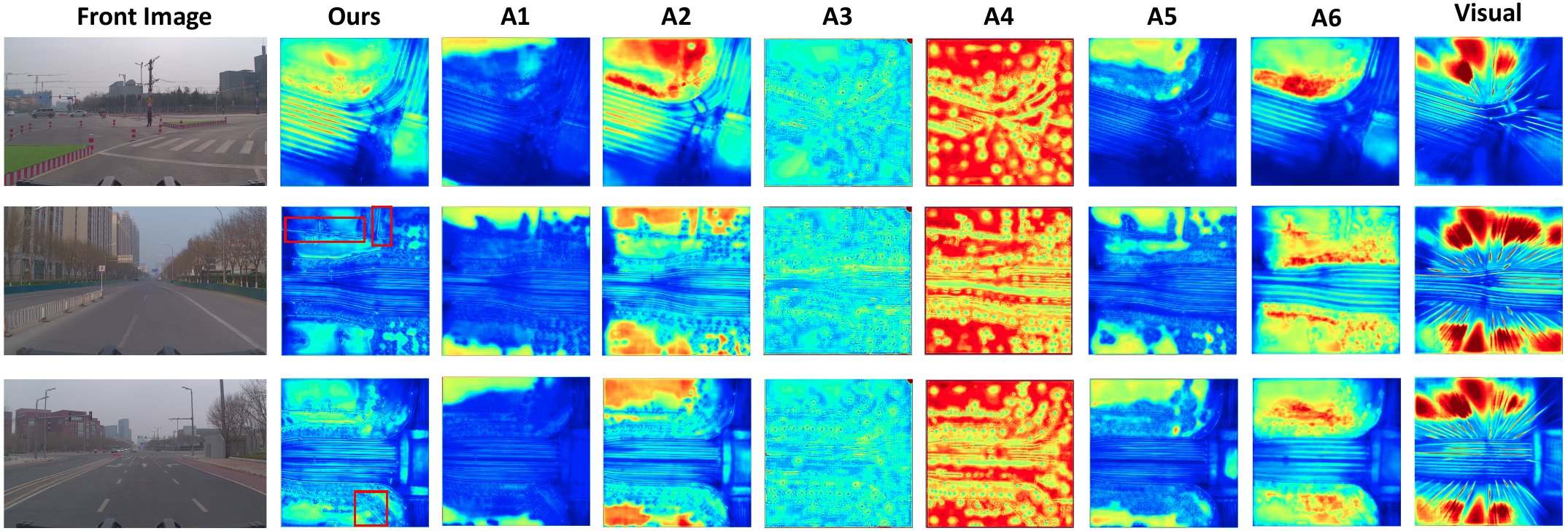}
	\vspace{-0.4cm}
    \caption{
        \small
		The three rows illustrate the visualization results of three distinct scenes. The leftmost column displays the front images of different scenes, and the remaining columns depict the BEV features of the ablation experiments and the two modes of our method.
    }
	\label{fig:long_graph}
	\vspace{-0.2cm}
\end{figure*}
\begin{table*}[h]
	\footnotesize 
	\begin{center}
		\scalebox{0.76}{
            \begin{tabular}{c|cccccc|ccc|ccc|ccc}
                \toprule
                \multirow{2}{*}{} & \multirow{2}{*}{\shortstack[1]{Surfel}} & \multirow{2}{*}{\shortstack[1]{Decoder}} & \multirow{2}{*}{\shortstack[1]{SemSup}} & \multirow{2}{*}{\shortstack[1]{SemEmb}} & \multirow{2}{*}{\shortstack[1]{$\mathcal{L}^{rp}_{KL}$}} & \multirow{2}{*}{\shortstack[1]{Histogram}} &
                \multicolumn{3}{c|}{Longitudinal Error} & \multicolumn{3}{c|}{Lateral Error} & \multicolumn{3}{c}{Yaw Error} \\
                &&&&&&& MAE(m) & RMSE(m) & $<$0.1m(\%) & MAE(m) & RMSE(m) & $<$0.1m(\%) & MAE($^{\circ}$) & RMSE($^{\circ}$) & $<$0.1$^{\circ}$(\%) \\
                \midrule
                Ours & \Checkmark   & \Checkmark   & \Checkmark   & \Checkmark   & \Checkmark & \Checkmark
                & \textbf{0.045} & \textbf{0.121} & \textbf{93.68}
                & \textbf{0.040} & \textbf{0.059} & \textbf{94.45}
                & \textbf{0.108} & \textbf{0.136} & \textbf{53.44} \\
                A1   & \XSolidBrush & \Checkmark   & \Checkmark   & \Checkmark   & \Checkmark   & \Checkmark
                & 0.092 & 0.298 & 90.09
                & 0.044 & 0.102 & 93.10
                & 0.109 & 0.144 & 53.36 \\
                A2   & \XSolidBrush & \XSolidBrush & \Checkmark   & \Checkmark   & \Checkmark   & \Checkmark
                & 0.093 & 0.317 & 83.63
                & 0.068 & 0.223 & 86.64
                & 0.115 & 0.176 & 52.80 \\
                A3   & \XSolidBrush & \Checkmark   & \XSolidBrush & \Checkmark   & \Checkmark   & \Checkmark
                & 0.094 & 0.345 & 89.23
                & 0.057 & 0.227 & 92.81
                & 0.123 & 0.168 & 49.83 \\
                A4   & \XSolidBrush & \Checkmark   & \XSolidBrush & \XSolidBrush & \Checkmark   & \Checkmark
                & 0.099 & 0.384 & 89.91
                & 0.050 & 0.157 & 92.80
                & 0.157 & 0.225 & 44.84 \\
                A5   & \XSolidBrush & \Checkmark   & \Checkmark   & \Checkmark   & \XSolidBrush & \Checkmark
                & 0.093 & 0.350 & 89.35
                & 0.049 & 0.132 & 91.74
                & 0.121 & 0.156 & 48.93 \\
                A6   & \XSolidBrush & \Checkmark   & \Checkmark   & \Checkmark   & \Checkmark   & \XSolidBrush
                & 0.116 & 0.352 & 77.70
                & 0.085 & 0.266 & 83.09
                & 0.161 & 0.310 & 45.58 \\
                \bottomrule
            \end{tabular}
        }
    \end{center}
    % \begin{tabular}{cccccccccccc}
    %     \hline
    %         Ablation & SemEmb & SemSup & Decoder & Surfel & Histogram & \makecell[c]{RMSE. \\ Lon.} & \makecell[c]{RMSE \\ Lat.} & \makecell[c]{RMSE \\ Yaw} & \makecell[c]{$<0.3$m \\ Lon. Pct.} & \makecell[c]{$<0.3$m \\ Lat. Pct.} & \makecell[c]{$<0.6^{\circ}$\\ Pct.} \\
    %     \hline
    %         Ours & \Checkmark & \Checkmark & \Checkmark & \Checkmark & \Checkmark & 0.082 & 0.065 & 0.229 & 99.70 & 99.97 & 99.79 \\
    %     \hline
    %         1st & \Checkmark & \Checkmark & \Checkmark & \XSolidBrush & \Checkmark & 0.148 & 0.074 & 0.237 & 99.27 & 99.91 & 99.76 \\
    %     \hline
    %         2nd & \Checkmark & \XSolidBrush & \Checkmark & \XSolidBrush & \Checkmark & 0.251 & 0.102 & 0.239 & 98.87 & 99.85 & 99.65 \\
    %     \hline
    %         3rd & \XSolidBrush & \XSolidBrush & \Checkmark & \XSolidBrush & \Checkmark & 0.176 & 0.080 & 0.248 & 99.37 & 99.89 & 99.62 \\
    %     \hline
    %         4th & \Checkmark & \Checkmark & \XSolidBrush & \XSolidBrush & \Checkmark & 0.176 & 0.088 & 0.238 & 98.89 & 99.78 & 99.49 \\
    %     \hline
    %         5th & \Checkmark & \Checkmark & \Checkmark & \Checkmark & \XSolidBrush \\
    %     \hline
    % \end{tabular}
    \vspace{-2mm}
    \caption{\small Ablation experiments of the key components of our method on the augmented self-collected dataset.}
    \label{tab:ablation}
    \vspace{-2mm}
\end{table*}
%\input{sources/tables/run_time.tex}

% shuang's version
%\vspace{0.1cm}
%\noindent
%\textbf{Map Size.}
%The map size comparison results are given in the Table.\ref{tab:map_size}, where the map sizes of MSF-LiDAR, DA4AD and our methods are 8.36MB/km, 5.92MB/km, and 0.35MB/km, respectively.
%Compared with MSF-LiDAR and DA4AD, the localization map size of our method decreases 95.8\% and 94.1\%, which validates the superiority of our approach.

% ChatGPT version & several round modifications
\vspace{0.1cm}
\noindent
\textbf{Map Size.}
Table \ref{tab:map_size} reports the map sizes of MSF-LiDAR, DA4AD and our method, which are 8.36MB/km, 5.92MB/km, and 0.35MB/km, respectively. 
Compared with MSF-LiDAR and DA4AD, our method achieves significant map size reduction by 95.8\% and 94.1\%, respectively, which demonstrates the compactness of our map.

%\vspace{0.1cm}
%\noindent
%\textbf{Run-time Analysis.}
%We evaluate the run-time performance of our method with a GTX 1080 Ti GPU, Core i7-9700K CPU and 16GB memory.
%The run-time of comparison methods are offered in the Table.\ref{tab:run_time}.
%The average and max processing time of our method are 120.31ms and 127.00ms, respectively.
%Although the average run-time of our method is longer than comparison methods, it already meets the requirement of real-time localization. Besides, the inference time of our method is more stable than MSF-LiDAR and DA4AD.

\subsection{Ablations}
% shuang's version
%To demonstrate the effectiveness of each module of our method, we conduct several ablation experiments on the self-collected dataset.
%In particular, in order to demonstrate the robustness of our method, we discard some kind of landmarks in the map with certain probabilities in the experiments of this section.
%The evaluation results are presented in the Table.\ref{tab:ablation}.

% ChatGPT version & modifications
To evaluate the effectiveness of each component of our method, we perform several ablation studies on our self-collected dataset. Specifically, to verify the robustness of our method, we randomly remove some types of landmarks from the map with different probabilities in this section. 
The results are shown in Table \ref{tab:ablation}.

%\vspace{0.1cm}
%\noindent
%\textbf{Surfel.}
%The ablation experiment A1 removes surfel landmarks.
%From the Table.\ref{tab:ablation}, it can be seen that the localization errors of A1 are larger than those of our full method, especially in the longitudinal direction.
%This is due to the fact that we randomly discard the pole elements in some frames, in which case our method with surfel suppresses the decline of longitudinal accuracy.
%It verifies the effectiveness of the extended surfel element proposed in this paper.

% ChatGPT version & modifications
\vspace{0.1cm}
\noindent
\textbf{Surfel.}
In experiment A1, we remove the surfel landmarks from the map. The errors of A1 are higher than those of our full mode, especially in the longitudinal direction. 
This is because after we randomly remove the pole features in some frames, the longitudinal constraints in these frames can only rely on the surfel features, and the removal of surfel features in A1 experiment will reduce the longitudinal localization accuracy. 
This demonstrates the effectiveness of the extended surfel features.

% shuang's version
%\vspace{0.1cm}
%\noindent
%\textbf{Transformer Decoder.}
%On the base of A1, the map embeddings are directly assigned with map queries by skipping the transformer decoder in the ablation experiment A2.
%Table.\ref{tab:ablation} shows that the localization errors of A2 are larger than those of A1, especially in the lateral direction.
%It demonstrates that the self-attention of map queries and cross-attention of map queries and BEV features in the transformer decoder are helpful to avoid mismatch.

% ChatGPT & modifications
\vspace{0.1cm}
\noindent
\textbf{Transformer Decoder.}
Based on A1, we remove the transformer decoder and directly use the map queries as the map embeddings in the experiment A2. 
The results show that A2 has larger errors than A1, especially in the lateral direction. 
This indicates that the self-attention of the map queries and the cross-attention of the map queries and the BEV features in the transformer decoder are crucial for our network.

\begin{table*}[h]
    \footnotesize
    \begin{center}
        \scalebox{0.82}{
            \begin{tabular}{c|c|ccc|ccc|ccc}
                \toprule
                \multirow{2}{*}{Method} &
                \multirow{2}{*}{AR} &
                \multicolumn{3}{c|}{Longitudinal Error} & \multicolumn{3}{c|}{Lateral Error} & \multicolumn{3}{c}{Yaw Error} \\
                && MAE(m) & RMSE(m) & $<$0.1m/0.2m/0.3m(\%) & MAE(m) & RMSE(m) & $<$0.1m/0.2m/0.3m(\%) & MAE($^{\circ}$) & RMSE($^{\circ}$) & $<$$0.1^{\circ}/0.3^{\circ}/0.6^{\circ}$(\%) \\

                \midrule
                GNSS-RTK
                & 93.67\%
                & 0.132 & 1.082 & 91.41/93.33/94.12
                & 0.122 & 0.950 & 91.26/93.63/94.54
                & - & - & - \\
                
                EgoVM
                & 99.82\%
                & 0.037 & 0.104 & 96.99/99.66/99.77
                & \textbf{0.026} & \textbf{0.035} & \textbf{98.54}/\textbf{99.93}/99.99
                & 0.131 & 0.506 & 42.49/96.29/\textbf{99.75} \\

                Fusion
                & \textbf{100.0\%}
                & \textbf{0.030} & \textbf{0.040} & \textbf{97.80}/\textbf{99.84}/\textbf{99.96}
                & \textbf{0.026} & 0.036 & 97.90/99.89/\textbf{100.0}
                & \textbf{0.122} & \textbf{0.192} & \textbf{44.41}/\textbf{97.31}/99.74 \\
				\bottomrule
            \end{tabular}
        }
    \end{center}
    \vspace{-2mm}
    \caption{\small Road testing evaluation of localization accuracy over 16,500 km in a test area with road network exceeding 1000 km.}
    \label{tab:road_test}
	\vspace{-2mm}
\end{table*}

\begin{table*}[h]
    \footnotesize
    \begin{center}
        \scalebox{0.83}{
            \begin{tabular}{cc|ccc|ccc|ccc}
                \toprule
                \multirow{2}{*}{Method} & \multirow{2}{*}{Scene} &
                \multicolumn{3}{c|}{Longitudinal Error} & \multicolumn{3}{c|}{Lateral Error} & \multicolumn{3}{c}{Yaw Error} \\
                && MAE(m) & RMSE(m) & $<$0.1m/0.2m/0.3m(\%) & MAE(m) & RMSE(m) & $<$0.1m/0.2m/0.3m(\%) & MAE($^{\circ}$) & RMSE($^{\circ}$) & $<$$0.1^{\circ}/0.3^{\circ}/0.6^{\circ}$(\%) \\

                \midrule
                \multirow{2}{*}{Ours (Visual)}
                & Day
                & 0.126 & 0.199 & 49.87/81.27/94.98
                & 0.057 & 0.077 & 83.92/98.44/99.54
                & 0.184 & 0.217 & 26.37/84.64/99.66 \\
                & Night
                & 0.160 & 0.365 & 46.95/79.06/91.80
                & 0.078 & 0.108 & 72.41/95.26/99.01
                & 0.238 & 0.301 & 24.61/69.46/96.49 \\

                \midrule
                \multirow{2}{*}{Ours}
                & Day
                & \textbf{0.032} & \textbf{0.041} & \textbf{97.95}/\textbf{100.0}/\textbf{100.0}
                & \textbf{0.029} & \textbf{0.036} & \textbf{99.38}/\textbf{100.0}/\textbf{100.0}
                & \textbf{0.088} & \textbf{0.114} & \textbf{66.44}/\textbf{97.88}/\textbf{100.0} \\
                & Night
                & 0.036 & 0.045 & 97.24/99.91/99.99
                & 0.032 & 0.051 & 97.39/99.25/99.49
                & 0.116 & 0.158 & 55.13/95.18/99.60 \\

				\bottomrule
            \end{tabular}
        }
    \end{center}
    \vspace{-2mm}
    \caption{\small Evaluation of localization accuracy for the two modes of our method on the day-night dataset.}
    \label{tab:night}
    \vspace{-3mm}
\end{table*}

% shuang's version
%\vspace{0.1cm}
%\noindent
%\textbf{Semantic Ablations.}
%On the base of A1, the ablation experiment A3 removes the semantic supervision (SemSup) and the A4 further replaces semantic embeddings (SemEmb) with semantic encodings as BEV-Lcator.
%We can see that compared with A1, the localization errors of A3 and A4 both increase, which validates the usefulness of the combination of semantic embeddings and semantic supervision.

% ChatGPT & modifications
\vspace{0.1cm}
\noindent
\textbf{Semantic Ablations.}
We remove the semantic supervision (SemSup) from A1 in the experiment A3, and we further replace the semantic embeddings (SemEmb) with semantic encodings as in BEV-Locator in the experiment A4. 
The results show that A3 and A4 both have larger errors than A1, particularly in the lateral direction. 
This shows that our network benefits from the combination of semantic embeddings and semantic supervision.

\vspace{0.1cm}
\noindent
\textbf{Random Pose KL Loss.}
The purpose of experiment A5 is to test the effectiveness of our proposed random pose KL loss $\mathcal{L}^{rp}_{KL}$. 
Compared with A1, the localization errors of A5 increase for all evaluation metrics, which shows that the random pose KL loss can improve the performance.
%\liangshuang{
%The aim of experiment A5 is to verify the function of proposed random pose KL loss $\mathcal{L}_{rp}^{KL}$.
%Note that the localization errors of A5 increase in terms of all evaluation criteria compared with A1, which demonstrates that the random pose KL loss is helpful for localization performance.
%}

% shuang's version
%\vspace{0.1cm}
%\noindent
%\textbf{Pose Solver.}
%In the ablation experiment A5, we replace our histogram-based pose solver with a regression-based one.
%The results in the Table.\ref{tab:ablation} present that the localization accuracy of regression-based method decreases compared with A1.
%Besides, regression-based pose solver is not interpretable and difficult to debug.
%Therefore, we choose the histogram-based pose solver in our method.

% ChatGPT & modifications
\vspace{0.1cm}
\noindent
\textbf{Pose Solver.}
We conduct the experiment A6 to examine the effect of the histogram-based pose solver, in which we use a regression-based pose solver instead of the histogram-based one in A1. 
The accuracy of the regression-based approach is lower than that of A1. 
Moreover, the regression-based pose solver lacks interpretability and is hard to debug. 
%Therefore, we prefer the histogram-based pose solver for our method.

% shuang's version
%\vspace{0.1cm}
%\noindent
%\textbf{BEV Feature Maps Visualization.}
%For a more intuitive understanding of the role of each module, we visualize the BEV feature maps of different modes and ablations of our method.
%In the Fig.\ref{fig:long_graph}, the leftmost column is the front image and the others are BEV feature maps of our various variant approaches.
%We can see that the BEV feature maps of ours, A1, A2, A5 and visual mode with semantic segmentation supervision show clear lanes, curbs, crosswalks and poles, while the BEV feature maps of A3 and A4 are blurry.
%It implies that the semantic supervision helps to generate high-quality and interpretable BEV feature maps.
%Besides, surfel features (e.g. buildings surfaces) are also clearly learned in the BEV feature maps of our full metohd marked by red rectangular boxes, which is helpful for precise localization.
%In the BEV feature maps of the visual mode, the positions of the pole elements are not accurately learned due to the inaccurate depth estimation, which explains the large longitudinal error of the visual method.

% ChatGPT version & modifications
\vspace{0.1cm}
\noindent
\textbf{BEV Feature Maps Visualization.}
To better understand the contribution of each key component, we visualize the BEV features of different modes and ablations of our method. 
Figure \ref{fig:long_graph} shows the front image and the BEV features. 
It can be observed that the BEV features of ours, A1, A2, A5, A6 and visual mode with semantic segmentation supervision have clear semantic elements such as lanes, curbs, crosswalks and poles, while the BEV features of A3 and A4 are blurry. 
This indicates that the semantic supervision enhances the quality and interpretability of the BEV features. 
Moreover, surfel features (e.g. building surfaces) are also distinctly learned in the BEV features of our full mode marked by red rectangular boxes. 
In contrast, the BEV features of the visual mode have inaccurate pole positions due to the erroneous depth estimation, which accounts for the large longitudinal error of the visual mode.

\subsection{Road Testing Evaluation}
%\liangshuang{
%The network was actually deployed on a fleet of RoboTaxi vehicles for road testing with a total mileage of 5,000-10,000km.
%The Table.\ref{tab:road_test} shows the localization accuracy performance of road testing of GNSS, our method and fusion method (ours and GNSS).
%The MAE of our method achieves 0.034m, 0.032m, and 0.329$^{\circ}$ in the longitudinal, lateral and yaw directions, which demonstrates the high accuracy of our method.
%Although the max localization error of our method in the longitudinal direction achieves 3.958m in view of the lack of landmarks that can provide longitudinal constraints in some degraded scenes such as underpass, the fusion method suppresses the max longitudinal localization error to 0.406m via integrating the measurements of GNSS.
%Consequently, our method keeps good localization accuracy in all road test scenarios, which is essential for fully unmanned autonomous driving.}
We build a multi-sensor fusion localization system based on error-state kalman filter (ESKF) that integrates EgoVM, GNSS-RTK, and inertial navigation, and deploy it to a fleet of RoboTaxi vehicles for road testing evaluation. 
%Our vehicles achieved zero disengagements from the localization system while running over 10,000 kilometers in a test area with mileage exceeding 1000 kilometers.
We tested our localization system in a complex urban area with over 1000 kilometers of road network.
By the time of paper submission, our system achieved zero disengagements in autonomous mode for more than 1500 kilometers. 
Moreover, we verified the robustness and accuracy of our system by running it in open-loop mode for over 15,000 kilometers before switching to autonomous mode.
Table \ref{tab:road_test} compares the localization accuracy of GNSS-RTK, EgoVM and fusion methods. 
We use four metrics: mean absolute error (MAE), root mean square error (RMSE), percentage of errors within certain thresholds, and available ratio (AR). 
AR is a new metric that measures the percentage of cases where the longitudinal, lateral and yaw errors are simultaneously less than 0.6m, 0.3m and 1$^{\circ}$ respectively. 
The fusion method achieves 100\% AR by combining the measurements of GNSS-RTK and inertial navigation, while EgoVM sometimes fails to provide accurate longitudinal localization due to the lack of landmarks in some degraded scenes.

%\liangshuang{
%Table \ref{tab:road_test} shows the localization accuracy performance of road testing of GNSS, our method and fusion method (ours and GNSS).
%The MAE of our method achieves 0.041m, 0.030m, and 0.127$^{\circ}$ in the longitudinal, lateral and yaw directions, which demonstrates the high accuracy of our method.
%In addition, we define a new evaluation metric termed as available ratio (AR), which represents the percentages of of longitudinal, lateral and yaw errors less than 0.6m, 0.3m and 1$^{\circ}$ simultaneously.
%Although the AR of our method is not 100\% in view of the lack of landmarks that can provide longitudinal constraints in some degraded scenes, the fusion method suppresses the max longitudinal localization error and reahces 100\% AR via integrating the measurements of GNSS.
%}

% shuang & guowei's version
%\subsection{Night Scenes Evaluation}
%We build another day-night dataset in the same roads to verify the effectiveness robustness of our method under low-light conditions.
%Specifically, we test the full mode and visual mode of our method in this dataset, where the localization errors are shown in the Table \ref{tab:night}.
%The localization accuracy of the visual mode in the night scenes is the worst.
%However, the full mode of our method still keeps high localization accuracy with assistance of LiDAR point clouds.

% ChatGPT & modifications
\subsection{Night Scenes Evaluation}
\begin{figure}[!htbp]
	\centering
	\includegraphics[width=1.0\linewidth]{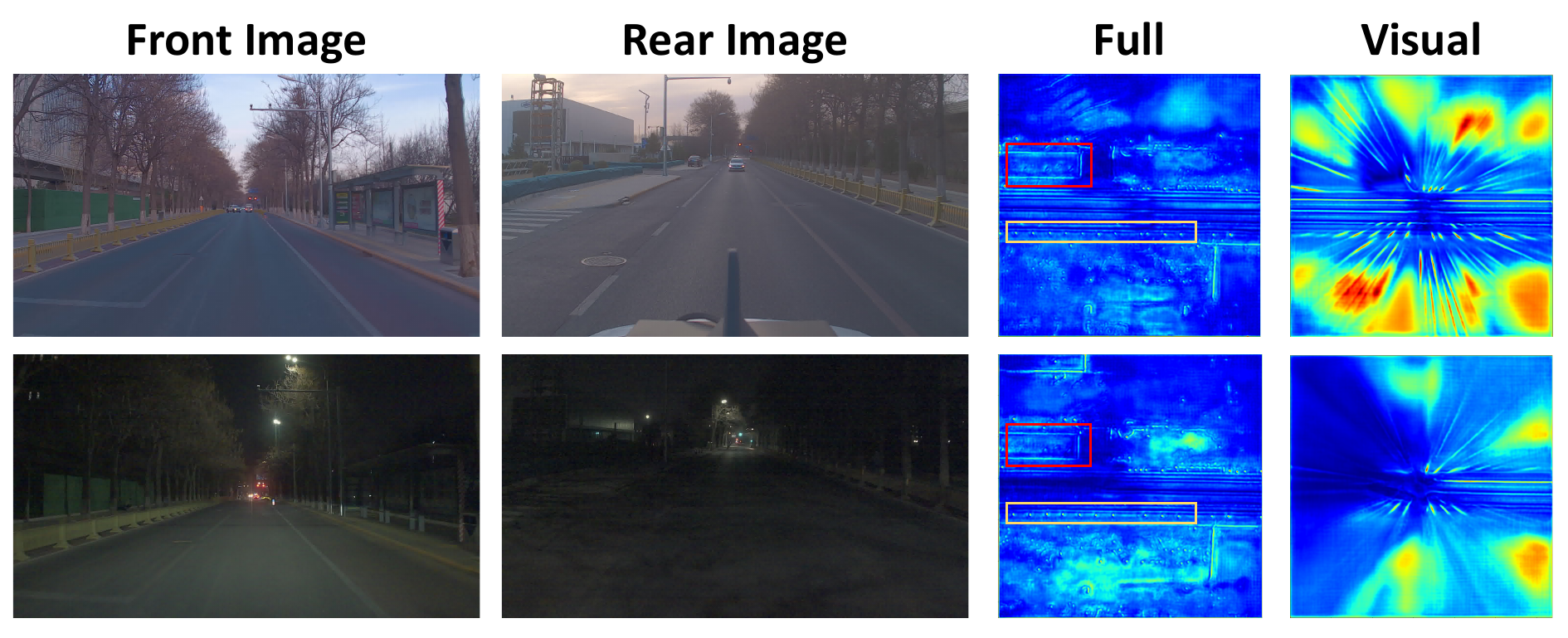}
	\vspace{-0.2cm}
    \caption{
        \small
		Visualization results on the day-night dataset. The first row displays the images and the BEV features of the full mode and visual mode of our method in a day scene. The second row shows the visualization results in a night scene at the same location.
    }
	\label{fig:short_graph}
	\vspace{-0.4cm}
\end{figure}

To test the robustness of our method under low-light conditions, we create another day-night dataset on the same roads. 
We report the localization errors of the full mode and visual mode of our method in Table \ref{tab:night}. 
The full mode achieves top performance with the aid of LiDAR point cloud, while the visual mode performs poorly in the night scenes due to the low visibility.
Figure \ref{fig:short_graph} shows the visualization results, where the first and second row display the day and night scenes, respectively. 
Each column shows the front image, the rear image, and the BEV features of the two modes, respectively. 
The front images reveal that the two scenes are at the same location with different illumination.
The BEV features of our full mode show clear lanes in the day scene but blurry lanes in the night scene, however, the poles and surfels are distinctly visible in both scenes marked by yellow and red boxes, ensuring the localization accuracy of our full mode. 
The BEV features of our visual mode show blurry lanes in the night scene and thus the localization accuracy deteriorates.

\vspace{-0.2cm}

%\vspace{0.1cm}
%\noindent
%\textbf{Day-Night Dataset Visualization.}
%Fig.\ref{fig:short_graph} shows the visualization results in the day-night dataset, where the first and second row presents the front image and BEV feature maps of full mode and visual mode of our method in a day and night scenario, respectively.
%From the front images, we note that the two scenes are almost at the same place, and the main difference is the light intensity.
%For the BEV feature maps of our full mode method (the second column), the lanes are clearly presented in the day scene but blurry in the night scene, however, the poles and surfels are apparently visible in both day and night scenes marked as yellow and red rectangular boxes, which guarantees the localization accuracy of our full mode method.
%For the BEV feature maps of our visual mode method (the third column), the lanes are blurry in the night scene and thus the localization accuracy decreases.
\section{Conclusion}
\label{section:conclusion}

We have proposed EgoVM, a novel end-to-end localization network that can improve localization accuracy to the centimeter level with lightweight vectorized maps in various challenging urban scenes.
We design a cross-modality matching module comprising learnable semantic embeddings supervised by semantic segmentation and a transformer decoder, which enhances the matching performance by transforming the two input modalities into a unified representation.
Moreover, we further improve the localization performance by incorporating LiDAR geometric features, which compensate for the deficiency of appearance features in certain scenes.
We have integrated our model with GNSS and IMU sensors to form a multi-sensor fusion localization system and have deployed it to a large fleet of autonomous vehicles, demonstrating its commercial viability.

{\small
\bibliographystyle{ieee_fullname}
\bibliography{egbib}
}

\newpage

%%%%%%%%% TITLE
Supplementary Material for ``EgoVM: Achieving Precise Ego-Localization using Lightweight Vectorized Maps"

\maketitle

\ificcvfinal\thispagestyle{empty}\fi

%\section{Map Elements Projection}

\setcounter{section}{0}
\setcounter{equation}{0}
\renewcommand\thesection{\Alph{section}}

\section{More Details about Network}
\noindent
\textbf{Network Configrations.}
We use VoVNet-39 as the image backbone in our experiments. The camera encoder takes images resized to $448\times 640$ as input and outputs a single-layer image feature map of size $28\times 40\times 256$. 
The LiDAR point cloud is cropped to $[-40m, 40m]$ in x-y dimensions and $[-3m, 5m]$ in z dimension before being fed into the LiDAR encoder, which produces LiDAR BEV feature map of size $160\times 160\times 256$ and resolution of $0.5m$. 
Both transformer decoders have four layers. 
The histogram search range in the pose solver is $[-3m, 3m]$ in x-y dimensions and $[-3^{\circ},3^{\circ}]$ in yaw dimension.

%The image backbone is implemented as VoVNet-39 in our experiments, and the input and output of camera encoder are images resized into $448\times 640$ and single layer image feature map with size of $28\times 40\times 256$.
%The LiDAR point clouds are cropped to the range of $[-40m,40m]$ in x and y dimensions and $[-3m,5m]$ in the z dimension and then fed into LiDAR encoder to obtain LiDAR BEV feature maps with size of $160\times 160\times 256$ and resolution of $0.5m$.
%The layers of two transformer decoders are both set as 4.
%The search range of histogram in the pose solver is $(-3m,3m)$ in x and y dimensions and $[-3^{\circ},3^{\circ}]$ in yaw dimension.

\vspace{0.1cm}
\noindent
\textbf{Multi-Layer BEV Feature Maps.}
%\tofix{Why single-layer is not enough and multi-layer is good.}
%\tofix{Describe each layer structure}
%\tofix{supervise three-layer bev feature maps}
%\tofix{semancit embedding projection}
As mentioned above, we set the resolution as $0.5m$ to guarantee the efficiency for extracting BEV feature map.
However, $0.5m$ is too coarse to achieve highly accurate localization, therefore, we upsampe the original BEV feature map twice to obtain finer BEV feature maps with resolutions of $0.25m$ and $0.125m$.
Meanwhile, we reduce the channel dimensions of the finer-layer BEV feature maps to increase efficiency.
Specifically, the sizes of three-layer feature maps $F^{B,0}$, $F^{B,1}$ and $F^{B,2}$ are $160\times 160\times 256$, $320\times 320\times 128$, $640\times 640\times 64$, respectively. 
In addition, we project original 256 dimensional semantic embeddings to lower dimensions by Conv1Ds and conduct semantic supervision for all three layer BEV feature maps using Equation 5.

%Correspondingly, we conduct semantic supervision for all three layer BEV feature maps.
%To adapt different channel dimensions of three layer features, we project original $256$ dimensional semantic embeddings to $128$ and $64$ dimensions by Conv2Ds with kernel of 1 and predict semantic probabilities for each layer feature map via Equation 5, respectively. 

%To achieve higher localization accuracy, we design multi-layer BEV feature maps and upgrade the semantic supervision and pose solver modules based on this.
%In our application, we upsample the original BEV feature map twice to obtain three-layer feature maps $F^{B,0}$, $F^{B,1}$ and $F^{B,2}$, whose resolutions and sizes are $0.5m$,$0.25m$,$0.125m$ and $160\times 160\times 256$, $320\times 320\times 128$, $640\times 640\times 64$, respectively.
%In semantic supervision module, the original semantic embeddings with $256$ dimensions are also projected by Conv2Ds with kernel of 1 to obtain semantic embeddings with dimensions of $128$ and $64$, and we conduct semantic supervision based on three-layer feature maps and semantic embeddings via Equation~(5).
%Moreover, in pose solver module, we also transform 256-dimension map embeddings to $128$ and $64$ dimensions via MLPs and then process multi-layer feature maps and map embeddings sequentially to refine localization results, as presented in the Algorithm \ref{alg:pose-solver}.

\vspace{0.1cm}
\noindent
\textbf{Multi-Level Pose Solver.}
%\tofix{unefficient to set fine grid size for [-3m,3m] and [-3deg,3deg], thus 0.5m grid size.}
%\tofix{however, grid size 0.5m not accurate}
%\toifx{Thus, the other two histograms with smaller search ranges and finer grid size.}
%\tofix{Details of each level histogram}
%\tofix{The combination of histogram, bev feature map and map embeddings.}
%\tofix{Alg 1}
To ensure efficiency, we need to select a proper grid size for the histogram, which has a large search range of $[-3m, 3m]$ in x-y dimensions and $[-3^{\circ},3^{\circ}]$ in yaw dimension. 
In practice, we set the grid size as $0.5m$, but this is too large to achieve highly accurate localization results. 
Therefore, we design an iterative multi-level pose solver that uses two additional histograms with smaller search ranges and grid sizes. 
The details of multi-level pose solver are described in Algorithm \ref{alg:pose-solver}.

%Due to the large search range of pose solver $[-3m, 3m]$ and $[-3^{\circ},3^{\circ}]$, the grid size of histogram should be selected properly to ensure efficiency.
%In practice, we set the grid size as $0.5m$, however, the grid size is too large to achieve highly accurate localization results.
%Thus, we add two complementary histograms by halving both search range and grid size twice and design an iterative multi-level pose solver to increase the localization accuracy as described in the Algorithm \ref{alg:pose-solver}.

%Moreover, in pose solver module, we also transform 256-dimension map embeddings to $128$ and $64$ dimensions via MLPs and then process multi-layer feature maps and map embeddings sequentially to refine localization results, as presented in the Algorithm \ref{alg:pose-solver}.

\begin{algorithm}[t]
    \caption{Multi-Level Pose Solver}
    \label{alg:pose-solver}
    \KwIn{BEV feature maps: $\{F^{B,l} | l\in\{0, 1, 2\}\}$;
		  Map embeddings: $\{M_i^{emb} | 1\leq i \leq K\}$;
		  Initial pose: $T_{init}$.}
    \KwOut{Estimated pose: $T_{est}$;\\
		   \ \ \ \ \ \ \ \ \ \ \ \ \ \ \ Estimated pose offset: $\Delta T_{est}$;\\
           \ \ \ \ \ \ \ \ \ \ \ \ \ \ \ Covariances: $\{\Sigma_l\}$.
		   }
    \BlankLine
    Initialize $T_{est}=T_{init}, \Delta T_{est}=0$ \;
    \For{$l\in\{0,1,2\}$}{
        Sample pose offsets by grid search, $\{\Delta T_{pqr}^l | 1\leq p,q,r \leq N_s\}\in \left [-\frac{r_x}{2^l},\frac{r_x}{2^l}\right ] \times \left[-\frac{r_y}{2^l},\frac{r_y}{2^l}\right] \times\left[-\frac{r_{yaw}}{2^l},\frac{r_{yaw}}{2^l}\right]$ ;
        
        $T_{pqr}^l = T_{est} \oplus \Delta T_{pqr}^l$ ;
        
        Unify BEV features and map embeddings, $F^{B,l} = Conv2d^l(F^{B,l})$,
		$ M_{i}^{{emb,l}} = MLP^l(M_i^{emb})$;
        
        Project map elements to BEV space, $p^{B}_i(T_{pqr}^l) = project(T_{pqr}^l, M_i)$;
        
        Sample features by bilinear interpolation,
        $M_i^{{bev}}(T_{pqr}^l)= bi\_inter(F^{B,l}, p^{B}_i(T_{pqr}^l))$;
        
        Obtain $\Delta T_l, \Sigma_l$ by equation (6), (7), (8);
        
        $T_{est} = T_{est} \oplus \Delta T_l$;
        
        $\Delta T_{est} = \Delta T_{est} + \Delta T_l$
    }
    Return $T_{est}$, $\Delta T_{est}, \{\Sigma_l | l\in\{0, 1, 2\} \}$
	\label{multi-lev}
\end{algorithm}

\begin{figure}[!htbp]
	\centering
	\includegraphics[width=0.8\linewidth]{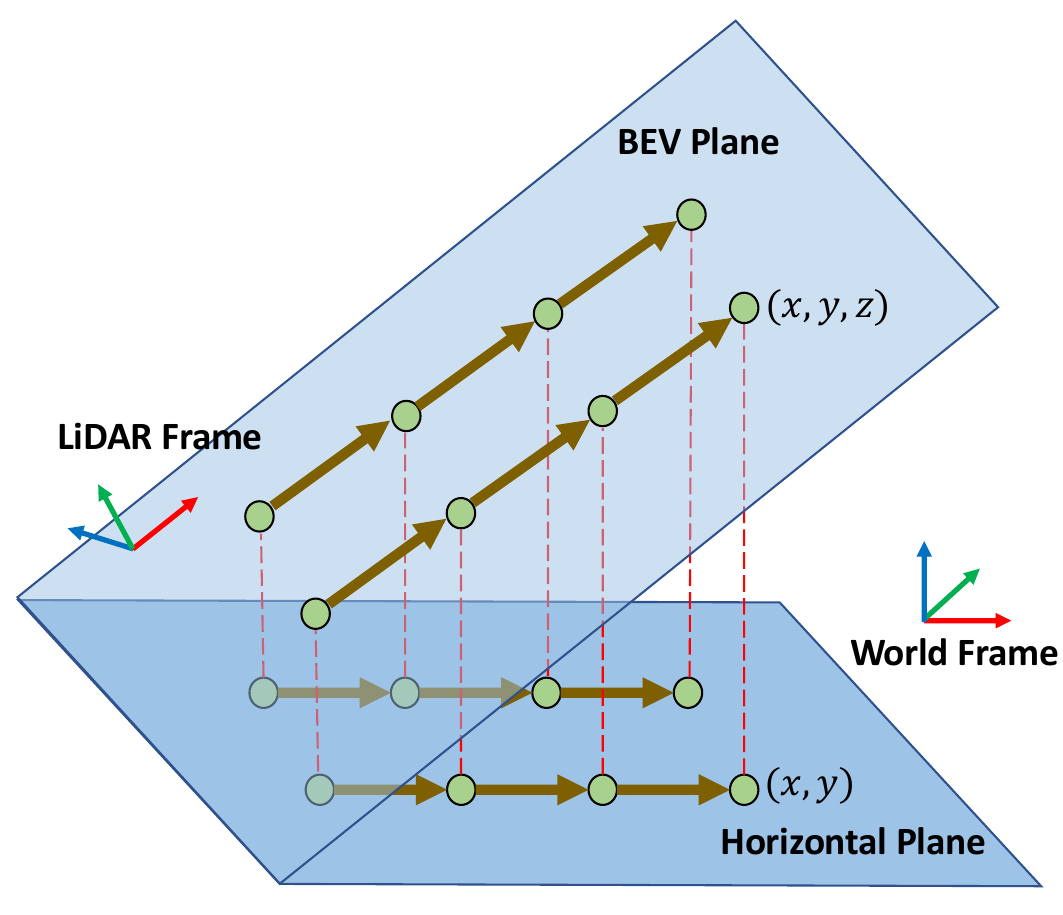}
	\vspace{-0.1cm}
    \caption{
        \small
        To project map elements, we first intersect vertical lines through the endpoints and the BEV space to get the z coordinates. 
        Then, we transform these 3D points into LiDAR frame and normalize them to obtain their representations in BEV space.
		%The visualizaiton of vectorized map elements projection. First, the z coordinates of the endpoints are calculated as the intersections of vertical lines through endpoints and the BEV space.
        %Then, these coordinates are projected into LiDAR frame and normalized to obtain their representations in the BEV space.
    }
	\label{fig:projection_graph}
	\vspace{-0.2cm}
\end{figure}

\noindent
\textbf{Map Elements Projection.}
BEV plane is affected by the rolling and pitching of vehicle platform, whereas vectorized map elements remain in horizontal plane.
To match map elements with BEV features, we need to transform map elements into the same coordinate as BEV features.
Firstly, we find the $z$ coordinate of the map element’s endpoint by intersecting the vertical line through $(x, y)$ and the BEV space, as illustrated in Figure \ref{fig:projection_graph}. 
The intersection of the 2D end point $(x,y)$ and the BEV space is denoted by $p=(x, y, z)$ and follows the equation below:
\begin{equation}
n^T(p - t_{WL})=0,
\end{equation}
where $n=R_{WL} (0,0,1)^T\triangleq (n_x,n_y,n_z)$ is the unit vector of z-axis of the LiDAR sensor in world frame and the value of $z$ is determined as follows:
\begin{equation}
    z = \frac{n^{T}t_{WL} - n_x x-n_y y}{n_z}
\end{equation}

%This section describes the process of HDMap elements projection from the world frame to the BEV space.
%Due to the fact that the endpoint $(x,y)$ of the vectorized map element in the HDMap does not store $z$ coordinate, we first calculate the $z$ coordinate as the intersection of vertical line at $(x, y)$ and the BEV space as illustrated in the Fig.\ref{fig:projection_graph}.
%Specifically, denote $p=(x, y, z)$ as the intersection of the 2D end point $(x,y)$ and the BEV space, which satisfies the following equation:
%\begin{equation}
%n^T(p - t_{WL})=0,
%\end{equation}
%where $n=R_{WL} (0,0,1)^T\triangleq (n_x,n_y,n_z)$ is the unit vector of z-axis of LiDAR in the world frame and the value of $z$ is determined as follows:
%\begin{equation}
%    z = \frac{n^{T}t_{WL} - n_x x-n_y y}{n_z}
%\end{equation}

Then we project $p$ from world frame to LiDAR frame:
\begin{equation}
    p^L={R_{WL}}^T  (p - t_{WL}).
\end{equation}

Finally, we normalize $p^L\triangleq (x^L, y^L,z^L)$ to obtain the coordinate of the end point $(x,y)$ in the BEV space:
\begin{equation}
    p^{B}=\left(\frac{x^{L}-h_{min}}{r}, \frac{y^{L}-w_{min}}{r}\right),
\end{equation}
where $h_{min}$ and $w_{min}$ are the minimum values in the BEV space, and $r$ is the resolution of each BEV grid.

%Then we project $p$ from the world frame to the LiDAR frame:
%\begin{equation}
%p^L={R_{WL}}^T  (p - t_{WL})
%\end{equation}
%Finally, we normalize $p^L\triangleq (x^L, y^L,z^L)$ to obtain the coordinate of end point $(x,y)$ in the BEV space:
%\begin{equation}
%p^{B}=\left(\frac{x^{L}-h_{min}}{r}, \frac{y^{L}-w_{min}}{r}\right),
%\end{equation}
%where $h_{min}$ and $w_{min}$ are the minimum values in the BEV space, and $r$ is the resolution of each BEV grid.

\section{More about Training}
\noindent
\textbf{Data Augmentation.} 
We use two data augmentation tricks in the training stage: LiDAR frame rotation and world frame rotation. 
For LiDAR frame rotation, we randomly rotate the LiDAR frame by $\theta$ around its z-axis and adjust the LiDAR pose, the LiDAR point cloud coordinates and the extrinsics from LiDAR to cameras accordingly. 
For world frame rotation, we randomly rotate the world frame by $\phi$ around its z-axis and adjust the LiDAR pose and the vectorized map element coordinates accordingly.

%In the training process, two data augmentation tricks are utilized in this paper: LiDAR frame rotation and world frame rotation.
%For the former, the LiDAR frame is rotated $\theta$ around its z-axis, thus the LiDAR pose, coordinates of LiDAR point cloud and extrinsics from LiDAR to cameras are changed accordingly.
%For the latter, we rotate the world frame $\phi$ around its z-axis, and then the LiDAR pose and coordinates of vectorized map elements are changed accordingly.

\vspace{0.1cm}
\noindent
\textbf{Training Strategy.}
We train our models on a NVIDIA A100 GPU cluster for 36 epochs with a batch size of 3 for each GPU. We set the base learning rate to $2\times 10^{-4}$. We use an image backbone pretrained on ImageNet with a learning rate multiplier of $0.1$ and a cosine annealing decay. We use AdamW optimizer with a weight decay of 0.01.

%Our models are trained on the A100 GPU with 36 epochs and batch size of 3.
%The basic learning rate is set as $2\times 10^{-4}$.
%The image backbone is pretrained on ImageNet, whose learning rate multiplier is $0.1$ and learning rate is decayed with a cosine annealing. We employ AdamW optimizer with weight deacy of 0.01.

\section{More about Ablations}

\vspace{0.1cm}
\noindent
\textbf{Augmented Self-collected Dataset.}
In order to test the significance of each module and increase the localization difficulty, we conduct ablation experiments by dropping some landmarks of the dataset. 
We use a random process to discard each type of landmark with a certain probability for each frame. 
Specifically, we set the drop probabilities for pole and curb landmarks to 5\% and 50\%, respectively.

%In the section of ablations, some landmarks of some frames in the dataset are dropped to increase the difficulty of localization so that the significance of each module can be fully verified.
%For each frame, each type of landmark is dropped with a certain probability.
%In our setting, the landmarks of pole and curb have 5\% and 50\% chances of being discarded for each frame, respectively.

\vspace{0.1cm}
\noindent
\textbf{Regression-based Pose Solver.}
In ablation experiment A5, we implement a regression-based pose solver instead of our histogram-based pose solver. 
The regression-based pose solver consists of a encoder block, a max-pooling layer and a MLP.
%We apply a encoder block, a max-pooling layer and a MLP to the map embeddings from the cross-modality matching module to get the predicted pose offset. 
The encoder block consists of three Conv1D(256,256,1) layers, where the frist two layers are followed by BatchNorm and ReLU.
The MLP consists of three 1D CNN layers: Conv1D(256,1024,1), Conv1D(1024,1024,1), Conv1D(1024,3,1), where the first two layers are followed by ReLU activation functions.

%In the ablation experiment A5, we replace the histogram-based pose solver with a reggresion-based one.
%Specifically, the map embeddings from cross-modality matching module are processed by a max-pooling layer and a MLP in sequence to obtain the predicted pose offset, where the MLP contains three-layer 1D CNNs: Conv1D(256,1024,1), Conv1D(1024,1024,1), Conv1D(1024,3,1) followed by ReLU non-linearity functions.

\section{More about Self-collected Dataset}
\noindent
\textbf{Sensor Suite Configuration.}
The sensor suite on our autonomous vehicle consists of a HESAI Pandar40P LiDAR, six FPD-LINK serial AR0231-AP0202-TI913 cameras and a NovAtel PwrPak7D-E1 integrated navigation system.

%The automatic driving cars in our self-collected dataset are equipped with a HESAI Pandar40P LiDAR, six FPD-LINK serial AR0231-AP0202-TI913 cameras and NovAtel PwrPak7D-E1 GNSS-IMU suits.
%In the self-collected dataset, we used the HESAITECH Pandor40P LiDAR, FPD\_LINK\_AR0231\_AP0202\_TI913 camera and NOVATEL PWRPAK7D\_E1 GNSS-IMU suits, respectively.

\vspace{0.1cm}
\noindent
\textbf{Ground Truth Poses Acquisition.}
Our goal is to obtain the ground truth poses of test frames that are consistent with vectorized maps. 
To achieve this, we first construct a large, precise and globally consistent point cloud map using point cloud registration techniques. 
Then, we generate vectorized maps based on this map.
Next, we use GNSS RTK/INS post-processing software like NovAtel Inertial Explorer to get the initial poses of test frames.
Finally, we align the test frames to the pre-built point cloud map using point cloud registration techniques again to get their ground truth poses. 

%To generate accurate and consistent ground truth poses, we first constructed a large, precise and globally consistent point cloud map via point cloud registration techniques.
%Then we used IEOUT post-processing software to obtain the initial pose of each frame and align the point cloud of each frame with the pre-built point cloud map to obtain the ground truth pose of each frame.
%In this way, the ground truth poses can be guaranteed to be consistent with the global map.

\section{Run-time Analysis}
We test the run-time performance of our method on a GeForce RTX 3080 GPU and a Intel(R) Xeon(R) W-3245 CPU @ 3.20GHz. 
The run-time statistics of comparison methods are shown in Table \ref{tab:run_time}. 
Our method has an average processing time of 120.31ms and a maximum processing time of 127.00ms. Although our method is slightly slower than others on average, it still meets the real-time localization requirement.
Moreover, our method has a more stable inference time than MSF-LiDAR and DA4AD. 

%We evaluate the run-time performance of our method with a GeForce RTX 3080 GPU and a Intel(R) Xeon(R) W-3245 CPU @ 3.20GHz.
%The run-time of comparison methods are offered in the Table.\ref{tab:run_time}.
%The average and max processing time of our method are 120.31ms and 127.00ms, respectively.
%Although the average run-time of our method is longer than comparison methods, it already meets the requirement of real-time localization. Besides, the inference time of our method is more stable than MSF-LiDAR and DA4AD.

\begin{table}[h]
    \small
    \begin{center}
        \resizebox{0.35\textwidth}{!}{
            \begin{tabular}{cccc}
                \toprule
                Method & Average(s) & 99\%(s) & Max(s) \\
                \midrule
                MSF-LiDAR & 51.91 & 111.76 & 316.00 \\
                DA4AD & 70.93 & 76.00 & 117.00 \\
                Ours & 120.31 & 125.00 & 127.00 \\
                \bottomrule
            \end{tabular}
        }
    \end{center}
    \vspace{-2mm}
    \caption{\small Run-time statistics of comparison methods.}
    \label{tab:run_time}
    \vspace{-4mm}

\end{table}

\end{document}